# AI-Driven Scholarly Peer Review via Persistent Workflow Prompting, Meta-Prompting, and Meta-Reasoning

Evgeny Markhasin
Lobachevsky State University of Nizhny Novgorod
https://orcid.org/0000-0002-7419-3605
https://linkedin.com/in/evgenymarkhasin

## Abstract

Critical peer review of scientific manuscripts presents a significant challenge for Large Language Models (LLMs), partly due to data limitations and the complexity of expert reasoning. This report introduces Persistent Workflow Prompting (PWP), a potentially broadly applicable prompt engineering methodology designed to bridge this gap using standard LLM chat interfaces (zero-code, no APIs). We present a proof-of-concept PWP prompt for the critical analysis of experimental chemistry manuscripts, featuring a hierarchical, modular architecture (structured via Markdown) that defines detailed analysis workflows. We develop this PWP prompt through iterative application of meta-prompting techniques and meta-reasoning aimed at systematically codifying expert review workflows, including tacit knowledge. Submitted once at the start of a session, this PWP prompt equips the LLM with persistent workflows triggered by subsequent queries, guiding modern reasoning LLMs through systematic, multimodal evaluations. Demonstrations show the PWP-guided LLM identifying major methodological flaws in a test case while mitigating LLM input bias and performing complex tasks, including distinguishing claims from evidence, integrating text/photo/figure analysis to infer parameters, executing quantitative feasibility checks, comparing estimates against claims, and assessing *a priori* plausibility. To ensure transparency and facilitate replication, we provide full prompts, detailed demonstration analyses, and logs of interactive chats as supplementary resources. Beyond the specific application, this work offers insights into the meta-development process itself, highlighting the potential of PWP, informed by detailed workflow formalization, to enable sophisticated analysis using readily available LLMs for complex scientific tasks.

## 1. Introduction

The rapid evolution of frontier large language models (LLMs) has significantly increased their power to handle complex expert-level tasks. This increasing power, in turn, stimulates research exploring ways to further expand LLMs' abilities and identify novel applications. Of particular interest are domain-specific STEM activities that continuously test human intelligence and push the boundaries of knowledge itself [1]. This focus is evident in the development of challenging benchmarks testing LLM abilities on problems ranging from international subject olympiads (e.g., OlympiadBench [2]) to graduate/expert-level STEM problems (GPQA [3], SuperGPQA [4], SciQA [5, 6], SciQAG [7], and Humanity's Last Exam [8]). At the same time, efforts are underway to develop LLMs with custom-tailored expertise and LLM-based expert systems [9–15]. Introduction of reasoning models, mimicking human thought process, constituted a significant advancement of general-purpose models' capabilities in the realm of complex tasks [16, 17], and this group of models is rapidly evolving [18–21]. While the capabilities of reasoning models like OpenAI o3 [22] and Google Gemini 2.5 Pro [23] represent significant advancements, these models remain limited when their training data lacks the specific domain facts or procedural knowledge necessary for devising effective solution workflows.

Several strategies can help bridge these gaps:

1. **Training a tailored model from scratch**: the most resource-intensive option, offering maximum control for specialized domains (e.g., protein chemistry) and tasks (e.g., chemical reaction extraction).
2. **Fine-tuning (adapting) existing models**: less resource-intensive than training from scratch but still requires domain-specific training data and expertise and faces certain constraints.
3. **Steering responses at inference time**: often the most practical approach that relies on advanced prompting techniques to provide necessary knowledge and workflow guidance directly within the prompt, requiring no changes to the underlying model and compatible with most available LLMs, including proprietary ones.

The third strategy generally relies on in-context learning (ICL [24–26]) and advanced prompt engineering techniques [27–35] to bridge the knowledge gap between model pre-training and the task at hand. Particular appeal of inference-time techniques stems from their ability to take full advantage of the most powerful frontier models, which incorporate

- the most expensive training (only accessible to select few vendors in the world),
- the best continuously improving world understanding,
- emerging multimodal analysis functionality,
- rapidly increasing inference-time limits.

Building on the potential of prompt engineering, this study focuses on developing and applying advanced prompting techniques to the challenge of AI-driven scholarly peer review, acting as a model complex problem of significant interest.

While this study uses AI-driven scholarly peer review as its primary complex application domain, the core prompt engineering methodology and the associated meta-development techniques (including meta-prompting and meta-reasoning used to formalize expert workflows) presented herein are intended for broader applicability across various complex analytical tasks. The specific proof-of-concept demonstration focuses on experimental chemistry manuscripts. Although advanced prompting may currently benefit from such domain specialization for achieving analytical depth, this focus serves primarily as a concrete testbed for developing and illustrating the general prompting architecture and meta-development principles that form the core contribution. Indeed, a key premise of this work is that this subject-agnostic development methodology can be readily adapted to engineer specialized prompts for peer-review-like analysis or other complex tasks in numerous other scientific and technical domains.

Consequently, while fully appreciating the nuances of the AI-generated chemistry analysis in the provided demonstrations might require some domain familiarity, understanding the prompt engineering methodology itself, the architectural concepts, the meta-reasoning examples, and the overall AI capabilities demonstrated should be accessible to the broader scientific and engineering community interested in advanced AI applications. Following the methodological discussions and examples requires primarily general scientific literacy, rather than deep expertise in the specific application domain chosen for illustration.

## 1.1. Scholarly Peer Review

Scholarly peer review is a cornerstone of academic research, demanding significant time, domain expertise, and critical reasoning. Using technical means to facilitate this process is a long-standing goal, which has gained urgency with the explosive growth of publishing activities and the recent advances in generative AI technologies increasingly used in academic publishing [36, 37]. The last few years alone have witnessed a wealth of publications addressing this automation problem via diverse approaches, including basic and methodological research [38–47], graph-based manuscript modeling [48], prompt-focused techniques [42, 46], probing capabilities of private and open-source models [38, 49, 50], investigations with reasoning models [40, 51], training custom models [50–52], developing multi-model/agentic systems [38, 48–50, 53, 54], and launching publicly accessible services [38, 43, 51, 55]. Due to its intellectually demanding nature, using AI for peer-review-like feedback also serves as a valuable method for evaluating and pushing the boundaries of advanced models.

Despite this progress, automating peer review remains a significant challenge for modern AI [39–41, 56]. Key difficulties include the inherent complexity of the task requiring deep, critical reasoning, the need for field-specific tailoring which involves capturing extensive tacit expert knowledge [57], and the historical lack of readily available, large-scale training datasets (with numerous attempts to address the latter issue [43, 45, 49–51, 58–65]).

Furthermore, existing approaches often face limitations. Training data consisting of high-level reviewer comments may not effectively teach models the detailed, step-by-step reasoning required for rigorous manuscript evaluation. Similarly, prompts based solely on common reviewer guideline questions (e.g., [66]) may fail to elicit the necessary depth of analysis compared to methods like chain-of-thought (CoT) prompting [67–69]. Critically, LLMs can exhibit inherent input biases, tending towards superficial agreement or outcome-based justifications rather than deep procedural scrutiny, further complicating the goal of objective evaluation. Addressing these interconnected challenges - codifying expert knowledge, designing robust analytical workflows, and actively countering cognitive biases - necessitates novel approaches.

## 1.2. Our Approach: Persistent Workflow Prompting

To address the limitations outlined above - specifically the need for detailed procedural guidance, the codification of tacit expert knowledge, and the mitigation of input biases - we explore an approach centered on advanced prompt engineering. Instead of relying solely on ICL examples or simple question lists, we focus on codifying the intellectual workflow inherent in rigorous peer review. Drawing inspiration from techniques like least-to-most prompting [70], task decomposition [71], plan-and-solve prompting [72], role-playing [73, 74], and PC-SubQ [75], we introduce Persistent Workflow Prompting (PWP). PWP utilizes a highly structured, hierarchical prompt that guides an LLM through a detailed analysis process designed to promote critical evaluation. This guidance involves decomposing the complex task of reviewing (specifically for experimental chemistry manuscripts in this work) into a sequence of manageable steps, effectively translating tacit expert knowledge [57] into an actionable protocol for the AI. This methodology aims to elicit deeper, more reliable analysis while counteracting default input bias tendency using only the standard chat interface of LLMs.

## 1.3. Scope and Limitations

Our investigation is deliberately constrained to using frontier LLMs accessible via standard chat interfaces, without relying on APIs, coding, or specialized tools, ensuring our methods are readily testable by a broad audience. Consequently, prompt engineering, specifically PWP, is the primary means of guiding the model. We focus on state-of-the-art reasoning models (primarily Gemini Advanced 2.5 Pro, also tested with ChatGPT Plus o1 [76] & o3, and SuperGrok Grok 3 Think) to maximize performance under these constraints. Key technical limitations influencing this work include model context window size, output token limits, and context recall accuracy [77, 78], which are particularly relevant given our goal of developing a large, structured prompt for analyzing full-length manuscripts and supporting information. While model limitations like hallucinations exist, their systematic characterization is beyond the scope of this initial study.

## 1.4. Contributions and Outline

While this paper details complex and abstract methodologies, it also provides readily accessible materials designed to facilitate understanding and quick replication via generally available AI chat bots. Key resources, including the full Markdown-formatted *PeerReviewPrompt* text for use with LLMs, a file with the *test paper* (including SI), usage protocol, and demonstration analyses are available in the Supporting Information, allowing readers to quickly test and verify the core PWP application described herein (**primary target model is Gemini Advanced 2.5 Pro**).

The main contributions of this paper are:

1. **Persistent Workflow Prompting (PWP):** We design, implement, and introduce PWP, a prompt engineering methodology employing a persistent, structured, workflow-based prompt to guide LLMs through complex, multi-step analytical tasks via standard chat interfaces.
2. **PWP Prompt for Chemistry Review:** We present a proof-of-concept PWP prompt specifically designed for the critical analysis of experimental chemistry manuscripts, demonstrating detailed workflow decomposition for this domain.
3. **Input Bias Mitigation:** We demonstrate that the developed *PeerReviewPrompt*, incorporating LLM's context conditioning via negatively biased persona engineering, can effectively counteract default positive input biases in LLMs, promoting a more critical and objective analysis of manuscript quality.
4. **Meta-Development Insights:** We describe the meta-prompting and meta-reasoning techniques used to iteratively develop and refine the PWP prompt, offering practical insights applicable to creating other complex, structured prompts.

5. **Empirical Demonstration:** We provide a qualitative demonstration and analysis of the PWP prompt's application using readily available reasoning LLMs, showcasing its ability to generate detailed, structured peer-review-like feedback incorporating multimodal analysis and quantitative checks.

The remainder of this paper is organized as follows:

- Section 2 details the methodology, including the meta-prompting techniques used (2.1), the architecture of the PWP prompt (2.2), and the process of formalizing the review workflow (2.3).
- Section 3 presents the results and discussion: Section 3.1 details highlights from the demonstration analyses; Section 3.2 discusses the rationalization and suppression of LLM input bias; Section 3.3 explores insights into PWP-guided LLM reasoning; Section 3.4 outlines study limitations; and Section 3.5 considers future directions.
- Supporting information provides basic PeerReviewPrompt usage protocol links to demonstration analyses (**direct links**: Gemini, ChatGPT chat, and ChatGPT analysis) and shared demonstration AI chats.
- Markdown-formatted prompt files for direct use with LLMs are included as PDF attachments and also shared via an OSF repository.
- Appendixes include the full text of the PeerReviewPrompt, demonstration analyses of the *test paper* (Gemini, ChatGPT, and Gemini - Baseline), and two sample prompts, supplementing discussions on applied and methodological meta-prompting.

Note: all examples with gray background contain **single** prompts or prompt templates (no AI responses), with some prompt examples consisting of several paragraphs or further structured with three horizontal dashes "---".

## 2. Methodology

### 2.1. Meta-Prompting

Meta-prompting represents an emerging methodology within prompt engineering [79, 80]. While a standard prompt typically targets a specific *actual problem* seeking a direct solution, a meta-prompt operates at a higher level of abstraction. It focuses on the prompting process itself, aiming either to refine the LLM's inference process for the actual problem or to generate a new prompt - the *Prompt Under Development* (PUD) - which will subsequently be used to solve the actual problem. This meta-prompting workflow thus often involves two distinct stages: **prompt generation** (developing the PUD) and **prompt execution** (using the PUD to address the actual problem).

The techniques encompassed by meta-prompting are valuable for refining even simple prompts and become indispensable when engineering complex prompts for challenging tasks. The prompt generation stage can utilize the same LLM intended for the subsequent prompt execution, a different model, or even specialized tools like Anthropic's prompt generator [81] with an XML-based meta-prompt. Given that prompt development itself can be complex, particularly for intricate target tasks, frontier reasoning models are often preferred for the prompt generation stage. For the complex problems addressed in this work, reasoning models were typically employed for both prompt generation and execution.

The meta-prompting techniques utilized in this work can be broadly classified into two groups based on their primary focus and how they engage the LLM:

1. **Linguistic and Structural Refinement:** This group includes techniques primarily aimed at improving the PUD text itself - its clarity, conciseness, grammar, and overall structure (akin to those described later in sections 2.1.1 and 2.1.2). In these approaches, the LLM generally functions as an advanced editor or proofreader, enhancing the prompt's readability and organization without necessarily analyzing the deep semantics of the instructions relative to the target task.
2. **Semantic and Workflow Engineering:** This group focuses on developing the functional core of the PUD, including its internal logic and detailed workflows (related to techniques in sections 2.1.4 and 2.1.5). A key characteristic here is that the meta-prompt or previous prompts within the chat often explicitly instruct the LLM to consider the *semantic meaning* of the PUD's content and to utilize the description or context of the *actual target task* when generating, refining, or validating the PUD's instructions (e.g., suggesting workflow steps). In this capacity, the LLM acts more as a collaborative partner or peer engineer, contributing directly to the design of the prompt's operational logic.

While there can be overlap, this distinction is useful. Developing sophisticated prompts like the *PeerReviewPrompt* typically requires applying techniques from both groups iteratively to ensure the prompt's language and structure are sound (Group 1) and to develop its complex workflows and logic using task-aware, semantically-focused meta-prompting (Group 2). Managing the complexity inherent in such advanced prompts necessitates careful structuring of the prompt text (using Markdown consistently in this study, edited primarily using Obsidian.md), benefiting both the human developer and the LLM's interpretation and facilitating the creation of hierarchical, modular prompts. The following subsections detail several specific meta-prompting techniques, illustrating these different approaches.

### 2.1.1. Language-Focused Refinement

One of the simplest meta-prompting approaches focuses directly on the linguistic quality of the PUD. In its basic form, the meta-prompt asks the LLM to improve the PUD text, for example:

```
Help me improve the following prompt:
---
{PUD Text}
```

This pattern primarily targets the linguistic and structural aspects of the `{PUD Text}`. By providing minimal guidance on *how* to improve it, the meta-prompt encourages the LLM to function like a human editor, applying general principles of clear technical writing, such as improving conciseness, grammar, and structure. However, unlike a human editor potentially unfamiliar with the subject matter, the LLM can leverage its "world knowledge" during meta-prompt processing to also consider and potentially refine the prompt's semantics.

More specific meta-prompts within this category can explicitly direct the LLM to focus on particular aspects, such as enhancing clarity, ensuring logical flow, or enforcing structural and grammatical parallelism (e.g., following principles outlined in AI-focused style guides like [82]).

### 2.1.2. Basic Iterative Refinement

Building on simple linguistic checks, a more interactive meta-prompting pattern involves soliciting feedback from the LLM to iteratively refine the PUD. The first step typically asks the LLM to analyze the PUD for potential issues:

```
Analyze the following prompt below (Prompt Under Development or PUD) and consider if
its instructions are clear, unambiguous, and complete. Provide feedback or ask
clarifying questions regarding any potential issues.
---
{PUD Text}
```

This pattern turns the meta-prompting process into a dialogue. Based on the LLM's feedback (e.g., questions about ambiguous instructions or missing details), the developer can provide clarifications. A subsequent meta-prompt then instructs the LLM to incorporate these clarifications and generate a revised PUD. The structure of this second meta-prompt can follow two strategies regarding the inclusion of the PUD text itself:

1. **Concise - Relying on Conversational Context**

The meta-prompt provides only the answers or clarifications, assuming the LLM retains the full PUD context from the previous turn:

```
Revise the PUD based on our previous discussion, incorporating the following
answers/clarifications. Analyze the revised prompt again: are there remaining
questions or unclear points? Provide additional feedback if necessary, or generate the
revised prompt with a clear, well-organized structure and precise language.

# Answers / Clarifications
{Developer's answers to LLM feedback}
```

- **Pros:** More concise input message. Often sufficient in continuous chat sessions with models exhibiting strong context recall.
- **Cons:** Susceptible to failure if the model's context window is exceeded, if context recall is imperfect (e.g., "lost in the middle"), or if the session is interrupted. Makes the revision step less self-contained and potentially harder to reproduce precisely.

2. **Verbose - Explicitly Providing Context**

The meta-prompt includes both the clarifications and the PUD text being revised:

```
Revise the prompt text provided below, incorporating the following
answers/clarifications. Analyze the revised prompt again: are there remaining
questions or unclear points? Provide additional feedback if necessary, or generate the
revised prompt with a clear, well-organized structure and precise language.

# Answers / Clarifications
{Developer's answers to LLM feedback}

---
# Prompt Text to Revise
{PUD Text - the version needing revision}
```

- **Pros:** More robust and explicit. Ensures the LLM operates on the correct PUD version, minimizing errors due to context loss. Each revision step is self-contained, aiding reproducibility and documentation. Recommended for very long/complex PUDs or when maximum reliability is needed.
- **Cons:** Results in a longer input message, which might seem redundant if context recall is perfect.

While we often employed Strategy 1 (concise) successfully during the highly interactive development phases described in this work, Strategy 2 (verbose) offers greater robustness, particularly for complex prompts or less predictable conversational contexts. This iterative cycle of feedback, clarification (using either strategy), and LLM-driven revision is a powerful technique for enhancing the clarity and effectiveness of complex prompts.

### 2.1.3. Meta-Meta-Prompting

As meta-prompts themselves become more complex and elaborate - potentially employing sophisticated prompt engineering techniques similar to those used in prompts targeting actual problems (like Anthropic's prompt generator meta-prompt [81]) - they too can benefit from LLM-based analysis and refinement. Applying meta-prompting techniques to improve another meta-prompt introduces a second layer of abstraction, a process termed *meta-meta-prompting*. In such a scenario, the LLM's output is not a PUD, but rather an improved meta-prompt intended for subsequent use in prompt generation (an example is shown in the initial part of chat [83]).

Given the higher level of abstraction (refining the prompt-generation tool rather than the problem-solving prompt), meta-meta-prompting often focuses primarily on the linguistic and structural refinement (Group 1 techniques) of the meta-prompt under development. Ensuring the meta-prompts instructions for the LLM generating the PUD are clear, well organized, and unambiguous is typically the main goal. However, for particularly complex meta-prompts that embed intricate logic or detailed workflow-generation procedures, applying semantic-focused techniques (Group 2) during meta-meta-prompting - analyzing the meaning and effectiveness of the meta-prompt's own instructions - can also be justified and beneficial. The DetailedMetaPrompt.md provided in the Supporting Information, designed for developing elaborate PUDs, serves as an example where such deeper refinement at the meta-meta level might be considered. Its application is demonstrated in a shared ChatGPT conversation [84]. This practice highlights how refining the tools used for prompt generation can be part of the overall development process for complex PUDs like the *PeerReviewPrompt*.

### 2.1.4. Workflow Generation and ICL-Based Techniques

This subsection explores several related meta-prompting techniques focused on generating or refining the core workflow within a PUD, often leveraging templates or examples.

1. **Template-Based Workflow Generation**

One approach uses a structured PUD template where the LLM is explicitly asked to devise the operational workflow. Consider this meta-prompt:

```
Analyze the following prompt template. Consider if the overall structure is clear.
Provide feedback/questions on any potential issues. Then, devise a detailed workflow
to replace the placeholder "{Workflow to be suggested by LLM}".

---
```

```
## Persona:
... (Description of a suitable role) ...

## Task:
... (Description of the task and task requirements) ...

## Processing Steps:
{Workflow to be suggested by LLM}
```

This technique leverages the LLM's ability to decompose complex tasks. Including such an explicit workflow often yields better PUD performance compared to relying solely on a high-level task description, even if the LLM could potentially infer the steps. The LLM-suggested workflow can then be reviewed and refined either manually or using iterative meta-prompting (Section 2.1.2). This pattern offers a balance between automated assistance and developer control, as illustrated in the development of the *modBibliographyHyperlinker* VBA module [85].

**2. ICL-Facilitated Prompt Generation**

In-context learning (ICL), typically used to provide examples of solving the *actual problem*, can also be applied during meta-prompting. Existing, well-structured prompts can serve as examples or references within a meta-prompt to guide the generation of a new PUD for a similar task:

```
Help me create a new prompt based on the reference prompt(s) provided below.
The new prompt should accomplish the following task:

## New Task Description
... (Description of the task for the new PUD) ...

---
# Reference Prompt(s)
{Full text of one or more existing prompts as examples}

---
Ask for clarification if needed before generating the new prompt.
```

Providing the reference prompts explicitly, as shown in the template above, ensures the LLM has the exact examples intended, analogous to the more robust verbose Strategy 2 discussed for iterative refinement (Section 2.1.2). However, similar to the concise Strategy 1 in iterative refinement, it is also possible to rely on the LLM's conversational context by simply asking it to use a specific prompt from earlier in the chat history as a reference, without explicitly resubmitting its text. This implicit approach is more concise but depends entirely on the model's ability to recall the relevant prior context accurately.

For instance, the development of the prompt for the *modZoteroFieldRecovery* VBA module successfully utilized this implicit context strategy, referencing the refined prompt for modBibliographyHyperlinker developed earlier within the same AI chat [86] without explicitly copying it. This example demonstrates the application of context-dependent ICL for prompt generation, though the explicit inclusion method offers greater reliability.

**3. Guided Workflow Generation**

While allowing the LLM to suggest a workflow from scratch (as in Technique 1 above) offers flexibility, it may sometimes lack sufficient direction for highly complex or nuanced tasks. In such cases, *Guided Workflow Generation* provides a more robust approach. Here, the developer manually creates an initial draft of the workflow - ranging from a high-level outline to a detailed protocol - and includes it within the PUD template given to the meta-prompt. The meta-prompt then asks the LLM to refine, complete, or elaborate on this provided draft workflow. This initial human guidance significantly constrains the LLM's output towards the desired structure and logic. This guided approach was fundamental to developing the detailed analysis protocols within the *PeerReviewPrompt* (specifically its Section IV) and was also used for the *MarkupProcessor* VBA module [83], where providing a detailed initial draft greatly simplified subsequent refinement.

### 2.1.5. Meta-Prompting for Complex Prompts

Developing highly complex PUDs often benefits from treating the LLM less like a simple tool and more like a collaborative partner or peer engineer, particularly when using state-of-the-art reasoning models. This approach involves more sophisticated meta-prompting techniques to refine intricate structures, logic, and content. Examples of such techniques used during the development of the *PeerReviewPrompt* include (see shared AI chats [87, 88] for further details):

- **Focused Refinement:** Targeting specific parts of the PUD for improvement.

```
Here is my current version of the prompt. Improve paragraph 1 in section D.2.
---
{Relevant excerpt of (or full) PUD Text, including section D.2}
```

- **Structure Optimization:** Collaboratively reasoning with the LLM about the PUD's high-level architecture, such as persona design or section organization, weighing pros and cons of different structures.

```
Do you see the two-level role of a researcher playing the role of a student? What
are the pros and cons of this architecture? If it doesn't provide obvious benefits,
how would I collapse it to a single role, while maintaining all features and
specifications related to the ultimate objective? [ … ]

Reflect on this idea: generating a collapsed single role per recommendations, but
reintroducing the student role not as a simulation, but specifying somehow this
behavior as part of the expert's role. [ … ]

Help me integrate the hybrid persona definition into the previous Expert Analyst
persona.
```

- **Reflective Refinement:** Asking the LLM to reflect on undesired behaviors observed during prompt execution and suggest improvements to the PUD to mitigate them.

```
How do I improve the original prompt to make sure you do not use reported results
for justifications [during analysis]?
```

- **Section Generation from Unstructured Notes:** Providing rough notes or questions and asking the LLM to structure them into a formal section of the PUD according to specified formatting rules.

```
Help me define section "5. A Priori Plausibility Assessment" from the following
text notes, formatting it as a bulleted list where each question/assessment becomes
its own bullet and all conditionals are dropped:
---
[Text notes, e.g., "Does the main result involve a process … superior compared to
existing alternatives … ? If yes, do authors identify … novel highly unintuitive
solution … ?"]
---
```

- **Reverse-Engineering with Generalization:** Analyzing a specific example of desired reasoning or output (potentially generated manually or in a separate context) and asking the LLM to generalize that process into abstract instructions suitable for inclusion in the PUD. For example, Section IV.D.2.3.F of the *PeerReviewPrompt* concerning quantitative feasibility checks was developed using this approach. First, the LLM was guided through a stepwise analysis of a specific process (similar to [89]). Then, the LLM was tasked with abstracting the reasoning from this specific analysis to create general instructions suitable for incorporation into the PUD. The objective was to generate instructions that would direct an LLM (when executing the updated PUD) to systematically identify suitable physical/chemical models, extract parameters, find governing equations, perform estimations, and compare with claimed results. A key constraint for this abstraction was to generate generalized instructions, avoiding terminology specific to the initial example analysis. (The source chat for this step was unfortunately lost.)

### 2.1.6. From Concept to Complex Prompt: Exploratory Meta-Prompting

Beyond refining existing prompt components, meta-prompting can structure the entire development life cycle for complex prompts, especially when building them from initial concepts or goals. This method establishes an interactive, multi-stage workflow where the LLM functions as a collaborator. Such an exploratory approach is particularly valuable for engineering prompts with intricate logic, tackling topics outside one's direct domain expertise, or tailoring prompts to specific applications or tasks, as the conversational process facilitates progressive refinement and discovery.

1. **Deep Research Prompts Development: Focus on Application**

This example illustrates general workflow by developing a deep research prompt for exploring a molecular biology topic "**Microplastics Interference with Mammalian Fertilization and Early Embryonic Development**". Being related to, but still outside of, the author's core expertise, this topic is convenient for demonstrating AI-assisted initial topic exploration combined with simultaneous development of a complex prompt for subsequent literature search. In fact, as the user engages in a conversation with AI, focusing on the actual subject, and not the prompt development process, the prompt engineering responsibility is largely shifted onto the AI. This shared AI chat [90] demonstrates the entire workflow, including the process of iterative refinement and diagnosing and correcting course when issues arise due to imperfect intermediate prompts or unexpected LLM responses. The final prompt is also included as an appendix.

A general strategy often involves the following stages:

a. **Seed Meta-Prompt:** Initiate the process with a simple meta-prompt stating the intent followed by initial, potentially rough ideas, questions, or areas of interest related to the research topic. It is acceptable for this seed prompt to use non-standard terminology or be linguistically unrefined, especially when exploring unfamiliar domains. For example:

```
Is there any research investigating distribution [sic] of the fertilization process
by micro plastic particles. The question to be answered include:
1) when micro plastic is found inside egg only, what are the chances of disruption
of parental DNA merging to form child DNA under semi natural conditions - the
process should still proceed in vitro, but the sperm needs to diffuse inside the
egg on its own.
2) When there are no detectable plastic particles inside the egg, but there are
such particles in sperm, what are the chances of particles getting inside the egg
under semi natural and artificial fertilization? What are the chances of subsequent
distribution [sic] of the first child DNA formation by plastic particles?
3) dependence of child DNA formation disruption on concentration of plastic
particles in sperm and egg under semi natural and artificial fertilization?
```

b. **Initial Revision:** Analyze the LLM's first response, paying close attention to how it interprets the initial ideas and any potentially ambiguous or non-standard terminology used in the seed prompt. Use a follow-up meta-prompt to provide clarifications and request that the LLM identify and replace informal terms with appropriate standard vocabulary (e.g., instructing the LLM to "refine the terminology" based on provided explanations). Performing preliminary research on terminology separately may not be necessary; the LLM can often assist directly with this terminology refinement based on context.
c. **Interactive Topic Development:** Engage in a natural, exploratory conversation with the LLM based on its responses and feedback. Discuss related concepts, refine existing questions, and explore new ideas or angles relevant to the intended research scope. This iterative dialogue helps progressively expand and focus the topic for the final deep research prompt.
d. **Generator Meta-Prompt:** Once the scope and key questions are sufficiently developed through interaction, use a "generator" meta-prompt to instruct the LLM to synthesize the discussion into a well-structured deep research prompt (the PUD). This generator prompt is often largely task-agnostic, focusing on requesting comprehensiveness, adherence to best practices for research prompts (e.g., including search terms, specifying output format like a detailed report with summary/abstract), and maintaining high academic standards. For example:

```
Ok, now generate a well-developed prompt, including exhaustive description of
discussed questions to be researched, search terms and phrases, etc. following best
practices for STEM focused deep research prompts, aiming to produce a detailed
report, meeting the highest academic standards. Make sure to call for creation of
report summary of abstract.
```

e. **Revision of Generated Prompt:** Critically examine the initial deep research prompt generated by the LLM. It is likely that further ideas, adjustments, or refinements will arise. Discuss these amendments iteratively with the LLM, instructing it to generate revised, standalone versions of the prompt (ensuring the final version does not contain conversational references like "based on our previous discussion" unless intended as part of the final prompt's structure).
f. **Final Deep Research Prompt:** The result of this iterative process is a well-developed, detailed prompt suitable for guiding an LLM in conducting the intended deep research task.

2. **AI Prompt Engineering Collaborator: Focus on Prompt Engineering Methodology**

This second example shifts the focus from using meta-prompting merely as a means to an end (developing a research prompt for another topic) to exploring AI-driven prompt engineering methodology itself as the subject. Here, the objective was to use a reflective, exploratory conversation with the LLM to surface prompt engineering capabilities implicitly available within the foundational model and then explicitly frame these capabilities within a sophisticated meta-prompt artifact. The full conversational log demonstrating the development of this meta-prompt is available via this shared AI chat [91].

Because this resulting meta-prompt (the "Adaptive Prompt Engineering Assistant & Tutor", provided in full in an appendix formatted for readability and as a Markdown source file for direct use with LLMs) is designed to assist in developing both regular prompts (PUDs) and other meta-prompts (mPUDs), it functions as what can be termed a meta$^2$-prompt. Consequently, the process of developing such a meta$^2$-prompt via interaction with an LLM formally represents meta$^3$-prompting - a third level of abstraction focused on creating versatile prompt-generation tools, representing a notable case of meaningful high-level abstraction in prompt engineering.

During this development process, the human user retains flexibility, allowing them either to delegate significant parts of the meta-prompt design to the AI or to participate more actively, providing specific guidance or minor steering. The final meta$^2$-prompt (or mmPUD) can be used subsequently to constrain or guide the LLM's capabilities when developing other, more specific prompts or meta-prompts. The core purpose of the "Adaptive Assistant" mmPUD developed in the demonstration is to configure an LLM to act as both an expert peer collaborator and an adaptive tutor, dynamically adjusting its interaction style.

## 2.2. Prompt Architecture: Hierarchical Modular Analysis Framework

### 2.2.1. Scope Definition and Development Test Case

Leveraging author's domain expertise in experimental physical chemistry, this field was selected as the target scope for the initial *PeerReviewPrompt* development. The prompt's detailed workflows and evaluation criteria were designed accordingly.

A crucial part of the iterative development process involved selecting a suitable test publication to serve as both a test case and a source of challenging analysis requirements. A specific publication [92] focusing on isotopic enrichment, known to contain significant and demonstrable methodological flaws, was chosen for this purpose. Its known issues made it a particularly informative test case for developing a prompt aimed at critical evaluation rather than simple summarization. The use of a single publication for development is a limitation of this initial proof-of-concept work, necessitated by resource constraints; testing on a broader range of manuscripts remains future work.

For practical testing during development, the input material used was the manuscript file combined with its corresponding supporting information (also see SI), taken exactly as provided by the publisher without structural modification or reformatting. This approach ensured testing occurred on realistic, commonly encountered input format.

### 2.2.2. Persistent Workflow Prompting (PWP)

The *PeerReviewPrompt* builds upon several advanced prompting techniques but introduces *Persistent Workflow Prompting* (PWP) as its core architectural principle. While incorporating standard top-level components like Role/Persona (Section II of the prompt), Context (Section III), and Task/Objective, the prompt's primary idea lies in its detailed, hierarchical structure designed to guide complex analysis. This structure moves beyond basic instructions to meticulously define *how* the analysis should be performed through explicit, multi-step workflows detailed primarily in the core Section IV. Specific Analysis Instructions.

The complexity of these workflows is managed using Markdown formatting within the prompt text (XML-based formatting is another potential alternative). This formatting is essential, serving both to organize the extensive instructions for the human developer/user and, critically, to aid the LLM in parsing and correctly interpreting the intended hierarchical structure and dependencies between different analysis steps.

The essence of PWP involves designing the initial, large prompt not merely as a single request, but as a *persistent workflow library* intended to be loaded into the LLM's context memory at the start of a session (the prompt explicitly states this intent in Sections III and V). Once loaded, this internal library of predefined workflows remains active. Subsequent, shorter user queries (e.g., "`Analyze the core experimental protocol`", "`Extract the main result`") act as triggers, invoking the relevant, detailed workflow(s) stored within the initial prompt's structure. This PWP approach enables complex, multi-turn analysis interactively without requiring the user to submit the large, detailed framework repeatedly, thereby preserving context window space for the manuscript and conversational history.

Section IV. Specific Analysis Instructions (Baseline Framework) of the *PeerReviewPrompt* serves as the primary workflow library. For instance:

- A query about the main result triggers the specific workflow defined in Section IV.B (Identifying Claimed Results and Contributions).
- A request to analyze a specific figure invokes the workflow detailed in Section IV.C (Analyzing Figures).
- A combined request like "`Analyze figures related to the main result`" prompts the model to chain workflows: first executing Section IV.B to identify relevant figures, then applying the Section IV.C workflow to each identified figure.
- Analyzing the core experimental protocol (covered in Section IV.D.2) involves prerequisite workflows (like those in IV.D.1, IV.B, and IV.C), executed logically based on overarching instructions (e.g., Section IV.A.3). (N.B.: The current implementation focuses core protocol analysis on key stages; full analysis requires further expansion.)

PWP activates this workflow library directly via the standard chat prompt, differentiating it from platform-specific features like Custom GPTs or Gemini Gems, which achieve persistence through separate mechanisms. The function of the PWP prompt thus extends beyond simple persistent instructions; it systematically encodes detailed procedures for complex analytical tasks, effectively acting as a high-level, declarative program written in natural language and structured using Markdown.

### 2.2.3. Behavioral Context and Persona Engineering

Beyond defining workflows, the *PeerReviewPrompt* utilizes Persona engineering within its Section II. Persona: Expert Critical Reviewer to instill specific behavioral characteristics crucial for critical analysis. While basic role prompting is common, this prompt employs a more elaborate approach. It explicitly rationalizes desirable traits of an expert reviewer (e.g., skepticism, objectivity, meticulousness) and attempts to project these traits onto the LLM through detailed role descriptions and associated expected behaviors.

To enhance the LLM's adherence to these traits, especially given the overall prompt's complexity and length, the persona definition is intricate, addressing multiple facets of the reviewer role. Furthermore, key behavioral instructions are deliberately repeated within the prompt architecture to mitigate potential issues arising from imperfect LLM context recall.

The primary goals driving this detailed persona engineering were:

- **Counteracting Outcome Bias:** A common failure mode observed was the LLM using reported results to justify the methodology. The persona instructions strongly and repeatedly emphasize a core principle of scientific review: methodology must be evaluated *independently* based on established scientific principles, irrespective of the claimed outcomes. A flawed method cannot produce reliable results, thus claimed results cannot validate the method itself.

- **Encouraging Analytical Rigor:** The persona aims to elicit detailed, critical, and well-justified output. It explicitly sets the expected standard of analysis as analogous to that required in a high-stakes academic examination (e.g., PhD qualifying exam), demanding meticulous attention to detail, clear reasoning, explicit detailed derivations, articulation of assumptions and arguments, and proactive identification of potential flaws or ambiguities.

A secondary aspect, addressed primarily within the main workflow instructions (Section IV.D.1.2) and reinforced by the persona, involves appropriately adjusting analytical expectations when evaluating proof-of-concept (PoC) studies. Such studies may feature certain deviations from strict scientific rigor, which can be acceptable if explicitly acknowledged and justified by the authors. This consideration is relevant even for the present work; this manuscript itself serves as a PoC relying on a single test case and qualitative evaluation without objective benchmarks, limitations, which are discussed further in Section 3.4.

### 2.2.4. Custom Classification for Guided Information Extraction

While LLMs can effectively extract specific information, such as a paper's main claimed result, interpreting this information for deeper, structured analysis may benefit from further guidance. Claims often intertwine distinct components:

- problem being addressed (the *unmet need*),
- proposed solution (methodology),
- specific *claimed novelty*.

A rigorous evaluation necessitates assessing these components independently - evaluating the problem's significance separately from the solution's validity and ingenuity.

To facilitate this analysis reliably, a custom classification scheme was developed and implemented in Section IV.B.1 (titled "Classification of the Main Claimed Result based on targeted unmet need"). This scheme provides the LLM with predefined categories relevant to experimental chemistry research. Its purpose is to guide the LLM, after identifying the main claim, to parse it into key components systematically by classifying the nature of the *unmet need*, the *proposed solution*, and *claimed novelty* according to these categories. For example, applying this scheme to the *test paper*'s [92] claim regarding "economical enrichment of $H_2^{17}O$... via slow evaporation and fractional distillation" guides the LLM to parse the claim into its core components: the *unmet need* (accessible $H_2^{17}O$), the *proposed solution methodology* ("slow evaporation and fractional distillation"), and any explicitly claimed novelty (importantly, allowing the scheme to guide the LLM in recognizing when, as in this case, specific novelty is not clearly articulated by the authors). This structured decomposition, guided by the custom classification scheme, enables a more consistent and rigorous downstream analysis (like the *A Priori* Plausibility Check described in Section 2.3.3) compared to relying on free-form claim interpretation.

### 2.2.5. Operational Guidelines and Default Procedures

The concluding section of the *PeerReviewPrompt* (Section V. Final Instructions for Interaction) establishes overall operational guidelines for the LLM's interaction and output. This final section serves several key functions aimed at ensuring consistent and high-quality analysis throughout a session:

- **Instructional Reinforcement:** This function involves strategically reiterating crucial directives presented earlier in the prompt. Specifically, this reinforcement includes emphasizing the core principles of the expert reviewer Persona (defined in Section II) and critical analytical constraints, such as the requirement for independent methodological scrutiny (from Section IV.A). Such repetition acts as a safeguard against context degradation or imperfect recall, promoting consistent application of the intended methodology.
- **Default Workflow Definition:** Here, the prompt specifies a "default" analysis workflow - a predefined sequence of analysis tasks (e.g., executing the comprehensive protocol analysis detailed in Section IV.D) - that the LLM should perform automatically when given a general, high-level request like "`Review this paper`". This default workflow provides a standardized and thorough starting point for analysis when specific sub-tasks are not initially requested by the user.
- **Output Formatting and Context Guidelines:** Finally, this section provides instructions regarding the format and context of the LLM's output. These guidelines can cover aspects such as structuring the response logically, using Markdown for readability, citing any external knowledge sources appropriately, and explicitly stating any assumptions made during the analysis. These output standards further ensure the generated review aligns with the rigorous expectations set by the Persona.

## 2.3. Formalizing the Review Process

### 2.3.1. Translating Expert Review into Actionable Prompts

A significant challenge in developing AI systems for tasks like scholarly peer review lies in translating the complex, often nuanced, reasoning processes of human experts into explicit, executable instructions suitable for an LLM. Expert review relies heavily on domain-specific knowledge, critical thinking, pattern recognition, and a considerable amount of tacit knowledge [57] - intuitions, heuristics, and ingrained understandings that experts apply subconsciously and often find difficult to articulate fully. Consequently, simply asking an LLM to "review a paper" typically yields superficial results, lacking the depth and critical rigor of true expert evaluation.

This limitation stems partly from the nature of generative pre-trained models. By default, LLMs often process input text by integrating it with their existing knowledge base, excelling at tasks like summarization where the input is largely taken at face value. Critical analysis, however, requires a different stance - one of abstraction and skepticism, where the input manuscript is evaluated against external principles and knowledge without being automatically accepted as truth. This critical stance, treating the manuscript as an object of scrutiny rather than incorporated fact, is generally not the default behavior and requires specific guidance. While frontier LLMs can perform complex abstract operations, eliciting in-depth critical analysis necessitates either specialized training or, as explored in this work, advanced prompting techniques designed to guide the model through a rigorous, structured evaluation process and overcome potential input biases (discussed further in Section 3.2).

The development of such advanced prompts for domain-specific expert-level analysis, exemplified by the experimental chemistry focus of this work's proof-of-concept (PoC), inevitably intersects with specialized knowledge and research practices inherent to any given field. While a deep dive into the specifics of, for instance, chemical knowledge formalization for AI applications in chemistry - areas covered by comprehensive reviews [93, 94] - is beyond the immediate scope of this methodology-oriented paper, such literature offers valuable context for domain-specific adaptations. Instead, our present work generally focuses on the more abstract framework development of PWP, which is an essential first step.

Therefore, creating the *PeerReviewPrompt*, focused on the experimental chemistry domain, necessitated a deliberate process of formalizing the intellectual workflow of critical review, aiming to make the implicit explicit and codify expert reasoning into a structured, actionable protocol. The subsequent sections detail this formalization process, including reflections on the meta-reasoning involved.

### 2.3.2. Deconstructing the Core Analysis Workflow

The process of formalizing the review workflow began by reflecting on how an expert typically approaches a manuscript. Rather than reading linearly like a novel, reviewers often seek specific high-level information first to orient themselves and determine the paper's core assertions.

The initial step usually involves identifying the *main claimed result* and any *key supporting findings*. This information is typically sought in the title, abstract, introduction, and conclusions. Understanding precisely *what* the authors claim to have achieved is paramount before evaluating *how* they claim to have achieved it. For the *test paper* [92], this step meant, for example, extracting the main claim about inexpensive $H_2{}^{17}O$ enrichment via specific methods. This task of identifying and extracting core claims was formalized into instructions detailed in Section IV.B (Identifying Claimed Results and Contributions) of the *PeerReviewPrompt*, including the custom classification scheme (discussed in Section 2.2.4) to help parse these claims structurally.

Once the core claims are understood, the focus shifts to evaluating the methodology described. A fundamental principle, emphasized throughout the *PeerReviewPrompt* (particularly in Section IV.A and the Persona definition) is the *independent assessment of methodology*. The validity of the experimental design, procedures, and data analysis must be judged based on scientific principles and best practices within the field, *without* relying on the claimed results as justification. This critical step involves scrutinizing the core experimental approach detailed by the authors, which corresponds to the analysis workflows initiated in Section IV.D (Analysis of Experimental Methodology) of the prompt. The goal at this stage is to determine if the described methods are fundamentally sound and capable, in principle, of supporting the types and magnitude of claims being made.

The first part of this methodology assessment, Section IV.D.1, defines a high-level workflow aimed at flagging obvious issues that may not require in-depth analysis. Beyond assessing general soundness, however, a deeper critical review, particularly in experimental sciences, often involves evaluating the *quantitative feasibility* of the claims based on the described procedures (formalized in Section IV.D.2.3.F). Simply stating that evaporation and distillation were used is insufficient; the reviewer must assess whether the *specific implementation* described could plausibly achieve the *magnitude* of the claimed result (e.g., the high isotopic enrichment reported in [92]). This assessment often requires comparing the reported outcomes against theoretical expectations derived from established scientific principles. This deeper analysis is guided by the workflows in the second part of the methodology section, IV.D.2, which focuses on the core methodology associated with the main claim.

The development of workflows for this in-depth methodological analysis (Section IV.D.2) was based on analyzing the test case [92]. Formal analysis of the paper's main claim involves theoretically estimating the performance of the described experimental setup using basic process models. To guide the LLM through such an analysis, Section IV.D.2 begins (after referencing Section IV.D.1 as a prerequisite) by directing the model to identify:

- The main claimed result and proposed solution / core methodology (via Section IV.D.2.1 and referenced Section IV.B.1).
- The individual experimental processes/stages comprising the core methodology (e.g., evaporation and fractional distillation in the *test paper*) via Section V.D.2.2.

This preliminary block identifies the target processes, allowing the LLM to be directed further to perform detailed analysis of *each identified stage* following the focused multi-step workflow defined in Section IV.D.2.3. To perform the quantitative analysis within this workflow aimed at assessing theoretical plausibility, the LLM must identify claimed quantitative characteristics for the stage while also determining suitable process models and collecting all parameters necessary for evaluating expected performance.

Section IV.D.2.3.A sets the stage by directing the model to extract preliminary stage-specific information and identify important components for further targeted extraction. Adding this structured information to the LLM's context aids subsequent workflow steps. The wording aims for generality across experimental chemistry while eliciting case-specific detail.

Section IV.D.2.3.B subsequently directs the LLM to identify and extract stage-related numeric quantities needed for theoretical modeling, while also incorporating instructions for initial identification of missing key information. Identifying missing information is crucial both for handling subsequent theoretical analysis (Section IV.D.2.3.F) and, potentially, for assessing the manuscript's completeness and the authors' awareness of limitations.

In experimental chemistry, equipment often plays a key role, addressed by the workflow in Section IV.D.2.3.D. In the test case, the fractionation column is key, but the paper lacks necessary details (e.g., theoretical plates, dimensions). When such details are missing, the prompt guides the LLM to meticulously analyze the manuscript and SI for clues. The *PeerReviewPrompt* includes a formalized missing information handling protocol distributed across sections IV.D.2.3.B, IV.D.2.3.C, and IV.D.2.3.D.

Section IV.D.2.3.C specifically directs the LLM to perform multimodal analysis of relevant figures (per Section IV.C description). This workflow was partly inspired by the test paper's flaws, particularly the lack of dimensions for the improvised distillation column shown in SI Fig. 1. The prompt guides the LLM to attempt scale estimation by identifying reference objects (like the 1 L flask mentioned in the text) visible in the photograph. The prompt language encourages detailed visual analysis, including extracting details not present in the text as a control against superficial processing.

While Section IV.D.2.3.E calls for a qualitative feasibility assessment, Section IV.D.2.3.F implements the core quantitative theoretical analysis using the following encoded steps:

1. **Select Appropriate Models/Equations:** Determine relevant physical/chemical models and governing equations for the identified process.
2. **Extract Explicit Parameters:** Gather necessary parameters explicitly provided in the manuscript/SI.
3. **Address Missing Information (Parameters):** Recognize missing critical parameters, involving:

    - *Inferring from Visual Data (Multimodal Analysis):* Analyze figures/diagrams/photos (leveraging prior analysis from IV.D.2.3.C).

- *Retrieving Standard Parameters:* Retrieve necessary fundamental constants or material properties.

4. **Perform Calculations:** Execute theoretical calculations using gathered parameters and models.
5. **Compare and Evaluate:** Compare estimated outcomes with claimed results to assess quantitative plausibility.

### 2.3.3. Formalizing Heuristics

Expert reviewers often develop intuitive heuristics or "rules of thumb" based on experience, which trigger skepticism or closer scrutiny even before detailed analysis. One such common heuristic is the "too good to be true" assessment. Formalizing such intuitive judgments into explicit prompt instructions is challenging but crucial for enabling deeper AI-driven critique. The process of developing the *A Priori* Plausibility Assessment (Section IV.D.2.5 of the *PeerReviewPrompt*) serves as a case study for this type of formalization, originating from the initial assessment that the claims in the *test paper* [92] seemed highly improbable.

Deconstructing this initial skeptical reaction involved identifying the underlying factors contributing to it. Reflection suggested the "too good to be true" assessment in this specific case arose from a combination of observations:

1. **Potentially Disruptive Claim:** The claimed result (cheap, accessible $H_2{}^{17}O$) promised significant impact, potentially disrupting existing markets (competing with commercially available expensive $H_2{}^{17}O$) and enabling new research avenues. High-impact claims often warrant higher scrutiny.
2. **Conventional Methodology:** The core experimental methods described (slow evaporation, fractional distillation) were well-established, widely understood, and generally considered conventional, lacking inherent novelty.
3. **Lack of Methodological Innovation:** The description of the experimental setup did not highlight any specific, non-trivial innovations or clever adaptations of the conventional methods that would plausibly explain the extraordinary outcome claimed. The apparatus appeared largely standard or even improvised (e.g., the packing material).
4. **Conflict with Established Knowledge:** Basic principles of physical chemistry suggest that achieving significant isotopic separation with the described simple methods and setup is extremely difficult, likely requiring far more sophisticated apparatus or processes.
5. **Absence of Author Justification:** The authors did not provide theoretical calculations, detailed process modeling, or other strong evidence within the paper to demonstrate the feasibility of their claimed results using the described methods, despite the apparent conflict with established knowledge (Point 4).

These factors were then translated into a structured set of questions and checks within (Section IV.D.2.5 of the prompt. This section guides the LLM to systematically assess the *a priori* plausibility by considering the scale of the claim versus the apparent novelty and sophistication of the methods, prompting for justification and checking for alignment with fundamental principles *before* delving into the quantitative verification of results. This formalization attempts to replicate the function of the expert heuristic by triggering targeted skepticism based on specific, identifiable characteristics of the claims and methodology presented.

### 2.3.4. Reflecting on the Knowledge Codification Process (Meta-Meta-Reasoning)

The process of developing the *PeerReviewPrompt* involved not only formalizing the steps of peer review (meta-reasoning) but also reflecting on *how* to effectively achieve that formalization, particularly when translating intuitive or tacit expert knowledge into explicit LLM instructions (*meta-meta-reasoning*). Attempting to codify one's own subconscious reasoning processes, as undertaken when deriving the quantitative checks (Section 2.3.2) or the plausibility heuristics (Section 2.3.3), presents unique challenges. This reflective phase aimed to identify potentially transferable strategies for developing complex, workflow-based prompts for other expert domains.

Several principles or strategies emerged from this meta-meta-reasoning process:

1. **Start with Concrete Cases:** Analyzing specific instances, like the demonstrably flawed *test paper* [92], provided concrete anchors for identifying both effective expert reasoning patterns and common pitfalls (like outcome bias) that the prompt needed to address or emulate.
2. **Trace the Reasoning Flow:** Consciously mapping out the sequence of questions, comparisons, and calculations an expert would likely perform (e.g., "What is the main claim?" -> "Is the method plausible?" -> "Do the numbers add up?") helped define the core structure of the prompt's workflows.

3. **Deconstruct Intuitive Judgments:** When faced with an intuitive reaction (e.g., "This seems too good to be true"), actively probing the basis for that intuition by asking "why?" and identifying the specific contributing factors (as detailed in Section 2.3.3) was key to translating it into objective, rule-based checks for the LLM.
4. **Isolate and Address Contradictions:** A primary goal of critical analysis is identifying inconsistencies. The formalization process focused on creating prompt instructions that explicitly direct the LLM to look for contradictions between:

- Claims and established scientific knowledge/principles.
- Claimed results and theoretical estimations based on the described methods.
- Different parts of the manuscript or supporting information.

5. **Identify Missing Information:** Recognizing that expert review often involves identifying crucial *omitted* details, the prompt development included steps specifically designed to check for and handle missing methodological information essential for validation or reproduction (as detailed in Section 2.3.2 - step 5) .
6. **Generalize Specific Analyses:** After analyzing a specific case, consciously abstracting the reasoning process and removing case-specific details was necessary to create generalizable workflow instructions applicable to a broader class of problems within the target domain (e.g., experimental chemistry papers). This step was crucial for the reverse-engineering technique mentioned in Section 2.1.5.
7. **Iterative Refinement (Linking back to Meta-Prompting):** The entire formalization process was not linear but iterative, relying heavily on the meta-prompting techniques (Section 2.1) to refine both the linguistic expression and the semantic logic of the prompt instructions based on trial runs and LLM feedback.
8. **Prioritize Sensitivity for Issue Flagging:** Recognizing the AI's role as an *assistant* primarily tasked with flagging potential issues for human evaluation, the design prioritized minimizing false negatives (missed issues) over minimizing false positives (incorrectly flagged issues). False negatives require laborious human rediscovery, while false positives can typically be dismissed more easily by the expert reviewer. Consequently, the prompt's persona (Section 2.2.3) and workflows were intentionally designed to encourage an *excessively critical* or *negatively biased* stance, aiming to maximize the identification of potential flaws, while treating the reduction of excessive false-positive "noise" as a secondary goal.

By consciously applying these strategies, it was possible to translate complex, often tacit, expert reasoning processes into the structured, explicit workflows embedded within the *PeerReviewPrompt*. These principles may offer guidance for others seeking to develop sophisticated prompts for complex analytical tasks in other domains.

### 2.3.5. Linking Formalized Procedures to PWP Architecture

The outcome of the formalization process described above - including deconstructed core analysis (2.3.2), formalized heuristics (2.3.3), and the overarching design principles (2.3.4) - directly informed the architecture and content of the *PeerReviewPrompt*.

Specifically, the detailed, multi-step procedures derived via meta-reasoning were implemented as the hierarchical workflows within Section IV (Specific Analysis Instructions) of the prompt. The Persistent Workflow Prompting architecture (Section 2.2.2) provided the mechanism to organize and store these complex, formalized procedures persistently within the LLM's context. Subsequent user queries then trigger these specific, pre-defined workflows, effectively guiding the LLM through the formalized expert reasoning process for tasks like analyzing the main result (Section IV.B), evaluating figures (Section IV.C), or assessing methodological plausibility and feasibility (Section IV.D). The persona engineering (Section 2.2.3) and operational guidelines (Section 2.2.5) further ensure that these workflows are executed with the desired critical stance and rigor.

In essence, the PWP architecture serves as the vehicle for delivering the formalized review process, translating the abstract principles and deconstructed steps identified through meta-reasoning (and further rationalized through meta-meta-reasoning) into an executable, natural language program for guiding advanced LLMs.

## 3. Results and Discussion

The *PeerReviewPrompt* was primarily developed using Google Gemini Advanced 2.5 Pro, with earlier exploration involving ChatGPT Plus o1. Demonstration analyses of the *test paper* [92] (including its SI) for several frontier reasoning models driven by this prompt are included in appendixes and linked in Supporting Information (Gemini Advanced 2.5 Pro, ChatGPT Plus o3, ChatGPT Plus o1 [95], and SuperGrok Grok 3 Think [96]).

As expected, the specific details and phrasing of the analyses varied between models and even between different runs on the same model. However, a key observation was the consistency in identifying core issues: all tested models, when guided by the *PeerReviewPrompt*, relatively reliably identified major methodological flaws within the *test paper* [92] and converged on the conclusion that its central claim (regarding isotopic enrichment) was highly dubious or unsupported by the described methods. This consistency across different architectures suggests the structured workflow provided by PWP effectively directs LLM reasoning towards critical evaluation points.

## 3.1. Highlights from Demonstration Analyses

A noteworthy aspect highlighted by the demonstrations relates to multimodal analysis capabilities. For example, Google Gemini Advanced 2.5 Pro (the subscription-based version) repeatedly demonstrated ability to analyze image content (specifically, photograph in SI Figure 1 of the *test paper* [92]) and integrate information extracted from visuals with the textual context, as guided by the PeerReviewPrompt. For instance, it consistently identified the presence of aluminum foil insulation around the fractionation column depicted - a detail absent from the main text. Furthermore, following prompt instructions, it successfully inferred approximate scale information from main text and applied this inferred data to subsequent steps involving the analysis of physical processes. While OpenAI has also indicated multimodal capabilities for its recent o3 reasoning model [22, 97], the limited testing performed during this work did not yield convincing evidence of integrated visual-textual analysis for this specific task. Furthermore, verifying the extent of such capabilities in ChatGPT models can be challenging due to the lack of transparency regarding their internal reasoning or step-by-step thought processes compared to models like Gemini Advanced.

Intriguingly, the LLM analyses highlighted at least two potentially significant issues not initially noted by the author during manual review. Firstly, multiple models consistently identified the use of a glass-wool-packed condenser as an improvised fractionating column as a poor methodological choice likely insufficient for the claimed separation. The models also usually suggested conventional accessible alternatives with potentially significantly higher and well-characterized performance. While evaluating this specific detail falls outside the author's direct expertise, the consensus across models and preliminary external checks suggest this criticism is likely valid. Secondly, several runs flagged inconsistencies related to the boiling points (b.p.) reported for different fractions (Table 1 in [92]). Although the prompt did not specifically target b.p. analysis (potentially explaining why this issue was not consistently flagged), the observation prompted closer scrutiny. Comparing the differences in reported uncorrected b.p. values between fractions reveals discrepancies when contrasted with known literature values for $H_2{}^{16}O$, $H_2{}^{17}O$, and $H_2{}^{18}O$ (which span only ~0.2°C at 1 atm according to [98], Table 9.1). This observation, combined with the authors' failure to monitor or report ambient pressure despite claiming a significant (10-15 times higher than the b.p. span of separated components) altitude-based b.p. depression for tap water, raises further critical questions regarding the meaning of the reported data and the entire study. This particular issue was initially missed by human review but surfaced by several PWP-guided LLM analysis runs.

These observations suggest the potential for PWP-guided LLMs not only to structure analysis but also to augment human review by identifying flaws that might be overlooked due to differing expertise or attention patterns. However, these findings are preliminary. A systematic comparison of analyses across models and multiple runs, potentially using quantitative metrics alongside qualitative assessment, is required for a rigorous evaluation of the prompt's performance, reliability, and limitations. Such a detailed comparative analysis was beyond the scope of this initial proof-of-concept study.

## 3.2. Input Bias: Rationalization and Suppression

A significant challenge in leveraging LLMs for critical tasks like manuscript evaluation is mitigating inherent reasoning biases. Early tests with simpler versions of the *PeerReviewPrompt* revealed a tendency for LLMs to exhibit what can be termed *positive input bias*. For instance, a model might identify potential flaws in an improvised experimental setup but still conclude the experiment was successful based solely on the manuscript's claim of achieving high enrichment (see example of naive analysis). Input bias is a known phenomenon, reported previously both in context of human [99] and LLM [100] reasoning. This observed behavior, where positive outcomes overshadow methodological scrutiny, can be also interpreted as *outcome bias* (as discussed in the context of persona engineering, Section 2.2.3). This tendency can be rationalized through the lens of modern LLMs' powerful In-Context Learning (ICL) capabilities.

ICL allows models to adapt and learn from the immediate context provided during interaction. This is evident in few-shot prompting, where models learn task patterns from examples, and more generally in conversation, where responses become progressively shaped by the preceding dialogue. The introduction of persistent memory further enhances this capability, enabling LLMs to incorporate knowledge from prior sessions, which helps compensate for training limitations like knowledge cutoffs. However, this very ability to learn from the input presents a fundamental tension: to learn effectively, the model must, to some extent, accept the provided material. This makes simultaneous critical evaluation - questioning the input's validity while also using it as learning material - an inherently difficult task requiring sophisticated abstract reasoning.

This challenge parallels aspects of learning in children, who often lack fully developed critical thinking skills and tend to accept learning materials as ground truth, taking them at face value without rigorous questioning (a point also relevant to translating expert review, Section 2.3.1). Our experience suggests the default operational mode of current frontier LLMs in chat interfaces often mirrors this, prioritizing contextual learning over inherent skepticism. Consequently, eliciting a genuinely critical treatment of input material typically necessitates deliberate, prompt-driven context conditioning.

Therefore, actively countering this default positive input bias was an essential goal in designing the *PeerReviewPrompt*. The strategies employed, including specific persona engineering elements (Section 2.2.3) and prompts designed to induce a more critical, negatively biased stance (2.3.4), aimed directly at shifting the LLM from passive acceptance towards active scrutiny. The demonstration analyses confirm that the negative bias conditioning implemented in the current *PeerReviewPrompt* successfully and reliably suppressed the observed positive input bias when applied to the *test paper* [92] using the target models, enabling a more rigorous evaluation.

## 3.3. Insights into PWP-Guided LLM Reasoning

The demonstrations highlighted the ability of LLMs, when guided by PWP prompt, to identify potentially significant methodological flaws, sometimes even those initially overlooked during manual human review (Section 3.1). Exploring how LLMs might generate such critical insights, despite lacking true scientific understanding or experimental experience, offers valuable perspectives on the interplay between their inherent capabilities and structured prompting methodologies like PWP.

Consider the example of multiple models identifying the use of a glass-wool-packed condenser as an improvised and likely inadequate fractionating column (reported in Section 3.1). One plausible mechanism behind flagging such issues relates to the LLM's fundamental nature as a predictive model trained on vast datasets. LLMs excel at learning statistical regularities and predicting text based on these patterns. Common and well-established scientific techniques, such as fractional distillation, are likely described extensively and relatively consistently within the LLM's training corpus, reflecting established scientific practices.

The PWP framework directs the LLM to extract specific methodological details from the input manuscript (e.g., "glass-wool packing" used for a "fractionation column"). The crucial step appears to be an implicit comparison: the LLM evaluates the likelihood of these extracted details from the paper against the patterns typically associated with the core concept (here, "fractionation column") derived from its general training data, which represents a form of encoded "world knowledge". If a specific detail from the paper (like "glass wool packing" in this context) corresponds to a low-probability pattern or significantly deviates from the common descriptions associated with standard fractionation columns found in the training data, the LLM may identify this as an anomaly, a potential inconsistency, or a contradiction to established practices.

This process arguably mirrors aspects of human scientific inquiry when encountering unfamiliar technical specifics. A scientist lacking deep expertise in fractional distillation might consult authoritative sources like textbooks or leading review articles to understand standard column designs. Alternatively, they might search scientific databases (like Google Scholar) to assess the prevalence and context of specific technical combinations, such as "glass wool packing" used for "fractionation columns". In scientific fields, frequently recurring methods and descriptions in reputable sources often reflect the prevailing scientific consensus on validity and best practice, analogous perhaps to the dominant patterns learned by the LLM.

The PWP methodology likely facilitates this type of pattern-comparison and anomaly detection. It does not teach the LLM chemistry, but rather structures the analytical process. By requiring the LLM to explicitly list experimental stages (e.g., Section IV.D.2.2), describe components in detail (Sections IV.D.2.3.D), evaluate feasibility (e.g., Sections IV.D.2.3.E, IV.D.2.3.F), and assess plausibility (e.g., Section IV.D.2.5), PWP focuses the LLM's pattern-matching capabilities on specific technical points for the purpose of critical evaluation. This structured approach, combined with the conditioning to mitigate input bias (discussed in Section 3.2), likely increases the probability of detecting and reporting low-likelihood details or inconsistencies compared to less constrained prompting methods.

Therefore, at least some of the critical insights generated by PWP-guided LLMs might arise not from deep reasoning in the human sense, but from a sophisticated comparison of manuscript-specific details against learned representations of scientific norms and consensus. Understanding this potential mechanism underscores the importance of detailed, structured workflows in prompting methodologies like PWP, as they serve to strategically harness and focus the LLM's powerful pattern-matching abilities for complex critical evaluation tasks.

## 3.4. Study Limitations

This study presents a proof-of-concept and, as such, has several important limitations that should be considered when interpreting the results and potential applicability of the PWP methodology:

1. **Single Test Case**: The *PeerReviewPrompt* was developed and primarily tested using a single publication [92]. Although chosen deliberately for its known flaws, this reliance on one test case limits the assessment of the prompt's generalizability to other experimental chemistry papers, particularly those that are methodologically sound or contain different types of errors.
2. **Limited Prompt Scope:** The current prompt workflows focus predominantly on the core experimental methodology described in the *test paper*, omitting rigorous analysis of crucial aspects like product characterization, subsequent experiments, subsidiary findings, data presentation, statistical validity, and introductory/concluding sections.
3. **Qualitative, Non-Benchmarked Evaluation:** The assessment of the prompt's performance presented herein is qualitative and observational. *No quantitative benchmark* was constructed for systematic evaluation against ground truth or objective metrics. Performance-related statements are based solely on the author's conventional (human-driven) evaluation of the generated LLM analyses, which introduces subjectivity and lacks comparison to defined baselines.
4. **Prompt Size and Platform Compatibility:** By design, PWP prompts incorporating detailed workflows can become very large (e.g., the *PeerReviewPrompt* exceeds 30 kB of text). While this complexity enables sophisticated guidance, the resulting size can exceed the input limits imposed by some widely available LLM chat interfaces. For instance, the official Qwen chat interface rejected the *PeerReviewPrompt*, citing its message input limit (10,000 characters, as of April 2025). This observation highlights a practical constraint, potentially restricting the direct applicability of large PWP prompts on certain platforms depending on their current input size limitations.
5. **Uncharacterized LLM Reliability:** While the PWP aims to guide LLMs towards rigorous analysis, inherent LLM limitations like potential hallucination or inconsistent context recall were observed occasionally but were not systematically characterized or quantified within this study. The impact of such issues on the reliability of the generated review feedback requires further investigation.

Collectively, these limitations underscore the preliminary nature of this work. Addressing them through broader testing, scope expansion, and systematic evaluation represents crucial future research directions.

## 3.5. Future Directions

The current *PeerReviewPrompt* serves as an initial proof-of-concept demonstrating the Persistent Workflow Prompting (PWP) methodology. Its development was intentionally focused on a limited scope (core experimental steps) and tested primarily against a single, deliberately chosen manuscript [92]. While this approach allowed for focused development and demonstrated the potential for PWP to guide complex critical analysis, several avenues for future work are apparent.

Key directions for further development include:

1. **Expanding the Test Set:** The most critical next step is to evaluate the current *PeerReviewPrompt* against a diverse set of experimental chemistry manuscripts, including those considered methodologically sound and those with different types of flaws than the initial test case. This is essential to assess the prompt's generalizability, identify its weaknesses, and guide further refinement.
2. **Broadening Analytical Scope:** The current *PeerReviewPrompt* workflows concentrate primarily on the core experimental protocol described for the main claimed result in the *test paper* [92] (i.e., the $H_2^{17}O$ enrichment via slow evaporation and fractional distillation). Significant expansion is necessary to apply similarly rigorous, workflow-guided analysis to other critical components typical of experimental papers, including aspects present in the test case itself that are not yet deeply scrutinized by the prompt. Key areas for scope expansion include developing workflows to evaluate:

- *Product Characterization Methods:* Critically assessing the techniques used to quantify or characterize the main product. For example, in the *test paper* [92], this analysis target would involve analyzing the GC-MS methods using 1-hexanol and hexamethyldisiloxane derivatives, the density and refractive index measurements, and the NMR analyses used to determine or verify enrichment.
- *Subsequent Syntheses/Applications:* Evaluating experiments where the primary product is used as a starting material. In the *test paper* [92], this analysis target includes the synthesis of $^{17}$O-labeled hydrogen peroxide via electrolysis and the preparation and characterization of $^{17}$O-labeled camphor.
- *Subsidiary Findings:* Analyzing the methodology, data, and claims related to secondary or unexpected results reported, such as the investigation into the camphor-catalyzed oxygen exchange reaction with ethanol described in the *test paper* [92].
- *General Manuscript Components:* Extending analysis beyond experimental procedures to cover data presentation (tables, figures beyond basic analysis), statistical validation (if applicable), the adequacy and clarity of the Introduction and Conclusions sections, and overall consistency throughout the manuscript.

3. **Optimizing Prompt Architecture:** While functional, the internal structure of the *PeerReviewPrompt*, especially the main workflow library (Section IV), warrants optimization based on insights gained during development and testing. All components should be reviewed for clarity, efficiency, and logical flow. Specific examples of potential architectural improvements include:

- *Streamlining Section IV.D (Methodology Analysis):* The current structure, including the detailed sub-steps within Section IV.D.2.3, could potentially be reorganized. For instance, consideration should be given to reordering specific analysis steps, such as performing the quantitative feasibility check (Section IV.D.2.3.F) *before* the qualitative assessment (Section IV.D.2.3.E), allowing the latter to incorporate the quantitative findings.
- *Adding New Checks:* The framework could be enhanced by adding checks for expected author-provided analyses. For example, a check could be added (perhaps also within Section IV.D.1.3) to verify whether the authors themselves performed and presented a *quantitative feasibility assessment*, especially for claims flagged as potentially unexpected or "too good to be true" by the *A Priori* Plausibility Assessment (Section IV.D.2.5).
- *Consolidating Related Checks:* Certain checks might be more logically placed within different workflow stages. For example, assessing whether authors explicitly reflected on or justified missing key experimental details could potentially be integrated into the General Red Flags section (Section IV.D.1.3).
- *Refining Claim Classification:* Given the distinct roles of the unmet need, the proposed methodology, and the claimed novelty identified during analysis (Section 2.2.4), the custom classification scheme could be refined**.** Future versions could involve developing separate, parallel classifications, for instance, to characterize the *type of solution methodology* employed (e.g., synthesis, separation) and the *nature of the claimed novelty* (e.g., new compound, improved efficiency), enabling more granular analysis.
- *Refining Triggering Logic:* The logic defined in sections like Section IV.A.3 that governs workflow execution based on user queries could likely be refined for robustness based on broader testing.

4. **Systematic Exploration of Multimodal Capabilities:** The preliminary success observed with Gemini Advanced 2.5 Pro integrating visual information highlights a promising research avenue. Future work should focus on systematically investigating and enhancing the use of multimodal inputs within the PWP framework. Prospective research directions include developing dedicated PWP workflows for more reliably extracting quantitative data from figures and tables, performing automated consistency checks between textual descriptions and visual representations (diagrams, graphs, photos), and potentially using visual data to assess the appropriateness or realism of described experimental setups. Evaluating the performance and limitations of such multimodal workflows across different capable LLMs is also an important target.
5. **Systematic Performance Evaluation:** A rigorous, systematic evaluation is needed that should involve comparing the outputs generated using PWP across different models and against baseline prompting techniques (e.g., zero-shot, simple role prompts) and, ideally, against actual human expert reviews, using both qualitative and quantitative metrics.

6. **Extending and Specializing PWP Applications:** The core Persistent Workflow Prompting (PWP) methodology appears potentially applicable to a range of complex analytical tasks beyond the current proof-of-concept. Future work could explore several avenues of extension and specialization:

- *Within Chemistry (Generalization and Specialization):* Beyond the current experimental focus, PWP could be adapted for theoretical chemistry manuscripts, requiring workflows tailored to evaluate theoretical frameworks, derivations, and computational methods. Furthermore, within both experimental and theoretical chemistry, opportunities exist for more specialized PWP designs targeting the specific nuances, common methodologies, and quality criteria of particular subfields (e.g., organic synthesis, analytical chemistry, quantum chemistry) or even the unique review standards of individual journals.
- *Peer Review in Other Sciences:* The PWP methodology could be tailored for scholarly peer review in other scientific disciplines (e.g., physics, biology, materials science, computer science) by collaborating with domain experts. For each discipline, similar to chemistry, both generalized PWP review prompts (e.g., for experimental biology) and more specialized versions targeting specific sub-disciplines or journals could likely be developed and prove useful.
- *Beyond Peer Review:* The PWP concept might also prove valuable for entirely different complex, multi-step analytical or procedural tasks outside of academic peer review, such as code generation with detailed control, detailed code review, analysis of laboratory notebooks, planning / designing experiments.

7. **Developing Advanced Benchmarking and Automated Refinement:** Systematic improvement requires robust evaluation methods. Future work should focus on creating specialized domain- or task-specific benchmarks designed for the complex STEM tasks targeted by PWP (e.g., critical chemistry review). Crucially, these benchmarks should incorporate evaluation protocols capable of assessing the performance not just overall, but also of individual modules or workflows within the hierarchical PWP structure - such as the module for main result extraction and classification (Section IV.B.1), the figure analysis workflow (Section IV.C), or even specific sub-steps within the methodology critique (e.g., the step for listing core experimental stages in Section IV.D.2.2) - enabling fine-grained diagnostics. Such detailed performance data could then be fed into LLM-driven meta-analysis. This step would involve designing novel structured meta-meta-prompts (potentially leveraging PWP principles themselves) to guide an LLM in analyzing performance patterns across different prompt sections against the benchmark results, thereby identifying specific areas for improvement. The ultimate goal is to establish a semi-automated or automated loop for semantics-driven prompt refinement, where benchmark data and LLM-based meta-analysis iteratively enhance the PWP prompt's effectiveness and reliability.
8. **Refining Meta-Development Processes:** Further research into the meta-prompting and meta-meta-reasoning techniques (Sections 2.1 and 2.3.4) could yield more systematic and efficient methods for developing complex workflow prompts like PWP.
9. **Investigating Foundational Model Capabilities:** Investigating foundational model development approaches (via training and/or fine-tuning) to intrinsically enhance critical evaluation capabilities (compensation of input biases) without diminishing robust ICL. Such model-level advancements are distinct from prompt engineering and lie beyond the scope of the present work and its anticipated continuation.

Addressing these points [points 1-8 specifically] will help mature the *PeerReviewPrompt* into a more robust tool and further validate the broader potential of the Persistent Workflow Prompting methodology, while recognizing that fundamental improvements in model capabilities [point 9] represent a separate, complementary research domain.

## 4. Conclusions

Eliciting deep, reliable, domain-specific reasoning from frontier Large Language Models (LLMs) using accessible methods remains a significant challenge, particularly for complex analytical tasks like critical scholarly peer review. This work addressed this challenge by introducing Persistent Workflow Prompting (PWP), a methodology centered on a detailed, hierarchical prompt acting as a persistent workflow library, developed through iterative meta-prompting and meta-reasoning designed to codify expert knowledge.

The proof-of-concept PeerReviewPrompt, targeting the critical analysis of experimental chemistry manuscripts, demonstrated this approach's viability. As qualitative demonstrations showed, the prompt successfully guided various frontier reasoning LLMs to perform complex, in-depth, and generally reproducible analyses of the test manuscript. Crucially, this guidance actively mitigated default input bias, enabling the reliable identification of major flaws within the prompt's defined scope while exhibiting robust performance across different models and runs. This outcome highlights the significance of the PWP approach. It showcases how sophisticated prompt engineering, informed by meta-reasoning, can translate expert workflows (including tacit knowledge) into structured instructions that actively condition the model for critical evaluation. This provides a feasible "zero-code" pathway to unlock specialized analytical capabilities within general-purpose LLMs, using only standard chat interfaces.

Looking ahead, further leveraging meta-reasoning and refining tacit knowledge codification should enable the development of PWP libraries. These libraries could guide LLMs through complex STEM problems (such as international olympiads or Humanity's Last Exam) using workflows similar to human experts. Furthermore, PWP-based approaches hold the potential to yield compatible performance across different frontier models and significantly improve the stability and reproducibility of solutions for complex, multi-step tasks. While the current work represents an initial demonstration requiring further validation and expansion, it underscores the power of structured, workflow-driven prompting as a key technique for advancing AI capabilities in demanding scientific and technical domains.

## Acknowledgments

Generative AI use has been an integral part of performed research, including interactive development of prompts via meta-prompting and extensive document revisions. This representative conversational log [101] documents the use of the Large Language Model Gemini (Google) to assist in the iterative revision and refinement of this manuscript. It serves as a demonstration of actively using AI as a peer collaborator during manuscript development. The documented interaction began with a draft manuscript that already included substantial preliminary revisions by the author and partial prior AI-driven revision.

## Supporting Information

## A. Prompt Files for Use with LLMs

**Note: the primary target model is Gemini Advanced 2.5 Pro.**

Prompt files are included as PDF attachments and are also available from:
https://osf.io/nq68y/files/osfstorage?view_only=fe29ffe96a8340329f3ebd660faedd43.

- *PeerReviewPrompt.md*: PWP-based experimental chemistry review prompt text for use with LLMs.
- *DetailedMetaPrompt.md*: Meta-prompt text for revision of prompts and meta-prompts (demo chat [84]).
- *Prompt_Engineer_Peer.md*: Prompt engineering assistant and tutor meta-prompt (demo chat [91]).

## B. Test Paper

To facilitate direct replication and review of the presented LLM analyses, the *test paper* (combined manuscript + SI [92]) PDF file used as input for the demonstrations is provided via a view-only link (Fair Use Statement): https://osf.io/nq68y/files/osfstorage?view_only=fe29ffe96a8340329f3ebd660faedd43.

## C. Demo Usage Protocol for *PeerReviewPrompt*

- **Message 1:** Input the full raw Markdown-formatted contents of *PeerReviewPrompt.md* in a new chat.
- **Message 2:** Submit "*Analyze the core experimental protocol*" prompt with the manuscript and SI attached.

Other sample prompts to try (manuscript only needs to be submitted once per chat):

- *Extract the main experimental result and key findings*
- *List all figures and tables directly associated with the core experimental protocol and main result*
- *Provide a detailed description of each figure associated with the core experimental protocol*

## D. Demonstration Analyses and Links

Included copies (appendixes) of demo analyses and full shared AI chats:

- Gemini - Critical Analysis of the Experimental Protocol for $H_2{}^{17}O$ Enrichment; shared AI chat [102].
- ChatGPT o3 - Core Experimental Protocol Analysis – Enrichment of $H_2{}^{17}O$; shared AI chat [103, 104].

Full shared AI chats only:

- ChatGPT Plus o1 [95]
- SuperGrok Grok 3 Think [96] (click on "Analysis of Core Experimental Protocol for $H_2{}^{17}O$ Enrichment" at the bottom).

Note: advanced features like modeling and multimodal analysis may yield variable or failed results.

## E. Shared Demo AI Chats

- Meta-prompting-based extended iterative prompt refinement [84] (see 2.1.3)
- Template-based and ICL-facilitated VBA module development [85, 105] (see 2.1.4)
- Guided workflow generation and VBA module development [85, 83] (see 2.1.4)
- Meta-prompting for *complex prompts* [87, 88] (see 2.1.5)
- Development of a deep research prompt [90] (see 2.1.6)
- Representative example of AI-driven workflow used for development of this manuscript [101].

H. Hoang, D.M. Cunningham, V. Goryachev, D. Patramanis, M. Krause, A. Redenti, D. Aldous, J. Lai, S. Coleman, J. Xu, S. Lee, I. Magoulas, S. Zhao, N. Tang, M.K. Cohen, O. Paradise, J.H. Kirchner, M. Ovchynnikov, J.O. Matos, A. Shenoy, M. Wang, Y. Nie, A. Sztyber-Betley, P. Faraboschi, R. Riblet, J. Crozier, S. Halasyamani, S. Verma, P. Joshi, E. Meril, Z. Ma, J. Andréoletti, R. Singhal, J. Platnick, V. Nevirkovets, L. Basler, A. Ivanov, S. Khoury, N. Gustafsson, M. Piccardo, H. Mostaghimi, Q. Chen, V. Singh, T.Q. Khánh, P. Rosu, H. Szlyk, Z. Brown, H. Narayan, A. Menezes, J. Roberts, W. Alley, K. Sun, A. Patel, M. Lamparth, A. Reuel, L. Xin, H. Xu, J. Loader, F. Martin, Z. Wang, A. Achilleos, T. Preu, T. Korbak, I. Bosio, F. Kazemi, Z. Chen, B. Bálint, E.J.Y. Lo, J. Wang, M.I.S. Nunes, J. Milbauer, M.S. Bari, Z. Wang, B. Ansarinejad, Y. Sun, S. Durand, H. Elgnainy, G. Douville, D. Tordera, G. Balabanian, H. Wolff, L. Kvistad, H. Milliron, A. Sakor, M. Eron, A.F.D. O, S. Shah, X. Zhou, F. Kamalov, S. Abdoli, T. Santens, S. Barkan, A. Tee, R. Zhang, A. Tomasiello, G.B.D. Luca, S.-Z. Looi, V.-K. Le, N. Kolt, J. Pan, E. Rodman, J. Drori, C.J. Fossum, N. Muennighoff, M. Jagota, R. Pradeep, H. Fan, J. Eicher, M. Chen, K. Thaman, W. Merrill, M. Firsching, C. Harris, S. Ciobâcă, J. Gross, R. Pandey, I. Gusev, A. Jones, S. Agnihotri, P. Zhelnov, M. Mofayezi, A. Piperski, D.K. Zhang, K. Dobarskyi, R. Leventov, I. Soroko, J. Duersch, V. Taamazyan, A. Ho, W. Ma, W. Held, R. Xian, A.R. Zebaze, M. Mohamed, J.N. Leser, M.X. Yuan, L. Yacar, J. Lengler, K. Olszewska, C.D. Fratta, E. Oliveira, J.W. Jackson, A. Zou, M. Chidambaram, T. Manik, H. Haffenden, D. Stander, A. Dasouqi, A. Shen, B. Golshani, D. Stap, E. Kretov, M. Uzhou, A.B. Zhidkovskaya, N. Winter, M.O. Rodriguez, R. Lauff, D. Wehr, C. Tang, Z. Hossain, S. Phillips, F. Samuele, F. Ekström, A. Hammon, O. Patel, F. Farhidi, G. Medley, F. Mohammadzadeh, M. Peñaflor, H. Kassahun, A. Friedrich, R.H. Perez, D. Pyda, T. Sakal, O. Dhamane, A.K. Mirabadi, E. Hallman, K. Okutsu, M. Battaglia, M. Maghsoudimehrabani, A. Amit, D. Hulbert, R. Pereira, S. Weber, Handoko, A. Peristyy, S. Malina, M. Mehkary, R. Aly, F. Reidegeld, A.-K. Dick, C. Friday, M. Singh, H. Shapourian, W. Kim, M. Costa, H. Gurdogan, H. Kumar, C. Ceconello, C. Zhuang, H. Park, M. Carroll, A.R. Tawfeek, S. Steinerberger, D. Aggarwal, M. Kirchhof, L. Dai, E. Kim, J. Ferret, J. Shah, Y. Wang, M. Yan, K. Burdzy, L. Zhang, A. Franca, D.T. Pham, K.Y. Loh, J. Robinson, A. Jackson, P. Giordano, P. Petersen, A. Cosma, J. Colino, C. White, J. Votava, V. Vinnikov, E. Delaney, P. Spelda, V. Stritecky, S.M. Shahid, J.-C. Mourrat, L. Vetoshkin, K. Sponselee, R. Bacho, Z.-X. Yong, F. de la Rosa, N. Cho, X. Li, G. Malod, O. Weller, G. Albani, L. Lang, J. Laurendeau, D. Kazakov, F. Adesanya, J. Portier, L. Hollom, V. Souza, Y.A. Zhou, J. Degorre, Y. Yalın, G.D. Obikoya, Rai, F. Bigi, M.C. Boscá, O. Shumar, K. Bacho, G. Recchia, M. Popescu, N. Shulga, N.M. Tanwie, T.C.H. Lux, B. Rank, C. Ni, M. Brooks, A. Yakimchyk, Huanxu, Liu, S. Cavalleri, O. Häggström, E. Verkama, J. Newbould, H. Gundlach, L. Brito-Santana, B. Amaro, V. Vajipey, R. Grover, T. Wang, Y. Kratish, W.-D. Li, S. Gopi, A. Caciolai, C.S. de Witt, P. Hernández-Cámara, E. Rodolà, J. Robins, D. Williamson, V. Cheng, B. Raynor, H. Qi, B. Segev, J. Fan, S. Martinson, E.Y. Wang, K. Hausknecht, M.P. Brenner, M. Mao, C. Demian, P. Kassani, X. Zhang, D. Avagian, E.J. Scipio, A. Ragoler, J. Tan, B. Sims, R. Plecnik, A. Kirtland, O.F. Bodur, D.P. Shinde, Y.C.L. Labrador, Z. Adoul, M. Zekry, A. Karakoc, T.C.B. Santos, S. Shamseldeen, L. Karim, A. Liakhovitskaia, N. Resman, N. Farina, J.C. Gonzalez, G. Maayan, E. Anderson, R.D.O. Pena, E. Kelley, H. Mariji, R. Pouriamanesh, W. Wu, R. Finocchio, I. Alarab, J. Cole, D. Ferreira, B. Johnson, M. Safdari, L. Dai, S. Arthornthurasuk, I.C. McAlister, A.J. Moyano, A. Pronin, J. Fan, A. Ramirez-Trinidad, Y. Malysheva, D. Pottmaier, O. Taheri, S. Stepanic, S. Perry, L. Askew, R.A.H. Rodríguez, A.M.R. Minissi, R. Lorena, K. Iyer, A.A. Fasiludeen, R. Clark, J. Ducey, M. Piza, M. Somrak, E. Vergo, J. Qin, B. Borbás, E. Chu, J. Lindsey, A. Jallon, I.M.J. McInnis, E. Chen, A. Semler, L. Gloor, T. Shah, M. Carauleanu, P. Lauer, T.Đ. Huy, H. Shahrtash, E. Duc, L. Lewark, A. Brown, S. Albanie, B. Weber, W.S. Vaz, P. Clavier, Y. Fan, G.P.R. e Silva, Long, Lian, M. Abramovitch, X. Jiang, S. Mendoza, M. Islam, J. Gonzalez, V. Mavroudis, J. Xu, P. Kumar, L.P. Goswami, D. Bugas, N. Heydari, F. Jeanplong, T. Jansen, A. Pinto, A. Apronti, A. Galal, N. Ze-An, A. Singh, T. Jiang, J. of A. Xavier, K.P. Agarwal, M. Berkani, G. Zhang, Z. Du, B.A. de O. Junior, D. Malishev, N. Remy, T.D. Hartman, T. Tarver, S. Mensah, G.A. Loume, W. Morak, F. Habibi, S. Hoback, W. Cai, J. Gimenez, R.G. Montecillo, J. Łucki, R. Campbell, A. Sharma, K. Meer, S. Gul, D.E. Gonzalez, X. Alapont, A. Hoover, G. Chhablani, F. Vargus, A. Agarwal, Y. Jiang, D. Patil, D. Outevsky, K.J. Scaria, R. Maheshwari, A. Dendane, P. Shukla, A. Cartwright, S. Bogdanov, N. Mündler, S. Möller, L. Arnaboldi, K. Thaman, M.R. Siddiqi, P. Saxena, H. Gupta, T. Fruhauff, G. Sherman, M. Vincze, S. Usawasutsakorn, D. Ler, A. Radhakrishnan, I. Enyekwe, S.M. Salauddin, J. Muzhen, A. Maksapetyan, V. Rossbach, C. Harjadi, M. Bahaloohoreh, C. Sparrow, J. Sidhu, S. Ali, S. Bian, J. Lai, E. Singer, J.L. Uro, G. Bateman, M. Sayed, A. Menshawy, D. Duclosel, D. Bezzi, Y. Jain, A. Aaron, M. Tiryakioglu, S. Siddh, K. Krenek, I.A. Shah, J. Jin, S. Creighton, D. Peskoff, Z. EL-Wasif, R.P. V, M. Richmond, J. McGowan, T. Patwardhan, H.-Y. Sun, T. Sun, N. Zubić, S. Sala, S. Ebert, J. Kaddour, M. Schottdorf, D. Wang, G. Petruzella, A. Meiburg, T. Medved, A. ElSheikh, S.A. Hebbar, L. Vaquero, X. Yang, J. Poulos, V. Zouhar, S. Bogdanik, M. Zhang, J. Sanz-Ros, D.

# A. PeerReviewPrompt Prompt

## A.1. Feature Highlights

- **Expert Peer Review Simulation**: Critically evaluates experimental methods before considering claimed results. Rigorously assesses protocols based on fundamental scientific principles to uncover hidden flaws and questionable assumptions, independent of claimed outcomes.
- **Information Extraction, Inference, and Integration**: Actively extracts crucial claims, numeric data, and procedural details from across the entire manuscript (text, tables, figures). Intelligently infers missing parameters and synthesizes disparate information with scientific knowledge to build a cohesive, evidence-based understanding.
- **Quantitative Reality Check**: Performs rapid back-of-the-envelope calculations, idealized modeling, and figure-based estimations. Rigorously tests if the described methods are quantitatively capable of achieving the reported results a priori, flagging claims potentially inconsistent with method simplicity.
- **Multimodal Figure Analysis**: Meticulously analyzes figures, photos, and schematics. Extracts quantitative details from visuals and cross-validates visual information against the text to uncover inconsistencies or provide unique supporting evidence.
- **Guided Analysis Framework**: Leverages in-context learning and a hierarchical, modular, and flexible prompt architecture that systematically guides the LLM through complex, multi-step critiques. Ensures thorough, consistent, and structured evaluation, acting like an interactive, expert-driven review template.
- **Zero-Code Accessibility**: Implements sophisticated manuscript analysis capabilities directly within the standard LLM chat window using generally available advanced reasoning models. Entirely prompt-driven, requiring no programming, API access, or specialized software installs. (Primary target platform - Gemini Advanced 2.5 Pro; also tested on Gemini Standard 2.5 Pro, ChatGPT Plus o1 and SuperGrok Grok 3 Think as of Apr 2025.)
- **Markdown-Driven Prompt Architecture**: Relies on inherent Markdown structure (headings, lists, bolding, MathJax extension) to organize complex instructions during the development process and to effectively guide the LLM's sophisticated analytical process within the chat interface. (Preserving format upon submission is essential for optimal function).

## A.2. Prompt: Critical Analysis of an Experimental Chemistry Manuscript

*Note: this is the main analysis prompt discussed in* Section 2.2

**WARNING:** This version is formatted for better readability.
It is not suitable for direct use with LLMs!
Use **PeerReviewPrompt.md** file from supporting information.

## I. Core Objective

Critically analyze the provided experimental chemistry manuscript (and any supporting materials) from the perspective of a highly skeptical expert. Identify potential flaws, inconsistencies, questionable methods, unsupported claims, or missing information, applying rigorous scientific scrutiny.

## II. Persona: Expert Critical Reviewer

**You ARE:**

1. **A Highly Qualified Chemist:** Possessing deep expertise in both experimental and theoretical chemistry, with broad academic and industrial research experience using diverse equipment.
2. **A Discerning Researcher:** You understand the differences between fundamental research, applied research, and proof-of-concept projects, tailoring your expectations accordingly.
3. **Critically Skeptical:** You **never** assume a manuscript is accurate, complete, or genuine, regardless of author, institution, or apparent publication status. Peer review can fail; data can be flawed, misinterpreted, or fabricated.
4. **Methodologically Rigorous:** You meticulously evaluate all aspects: theory, setup, protocols, data, assumptions, calculations, and conclusions. You demand robust evidence, especially for novel or unexpected findings.
5. **Practically Aware:** You recognize that non-conventional choices (equipment, procedures) occur but **require strong scientific justification**. You assess:
   - **Rationale vs. Rigor:** Is the choice justified by necessity (cost, availability, specific goal) or merely convenience? Does it compromise essential aspects for the research stage (e.g., a shortcut acceptable for PoC might be unacceptable for validation)?
   - **Performance Impact:** Could the choice negatively affect key metrics? Can meaningful results still be obtained? Is a standard, accessible alternative clearly superior?
   - **Validation Complexity:** Does the non-conventional choice complicate the interpretation or verification of results, *especially* if they are unexpected?

**Your Mandate & Performance Standard:** Maintain the highest standards of scientific integrity. Challenge assumptions, verify claims, and ensure conclusions are unassailably supported by the evidence presented *and* established chemical principles. **Execute this critical analysis with the performance standard expected during a high-stakes academic evaluation (such as a PhD or postdoctoral qualifying exam):**

- Embody meticulous rigor.
- Complete transparency in your reasoning.
- Explicitly show all calculation steps and assumptions.
- Identify and reflect on missing essential information.
- Actively look for inconsistencies, ambiguities, alternative interpretations, logical fallacies, impossible claims, or data that contradicts known principles.
- Demonstrate strict adherence to the analytical instructions provided.

## III. Context: Framework for Critical Manuscript Review

This prompt establishes a framework for conducting **in-depth, critical reviews of experimental chemistry manuscripts**. Your assigned **Persona** (Section II) defines the expert perspective, skeptical mindset, and high performance standards required - mirroring the rigor expected in demanding academic or industrial evaluations.

The **Specific Analysis Instructions** (Section IV) detail distinct methodologies and analytical checklists (e.g., for figures, protocols). Consider these instructions as a **structured toolkit** designed to guide your critique.

**How to Use This Framework:**

1. **Persistent Foundation:** This entire prompt (Persona, Context, Instructions, Final Rules) serves as the foundation for our entire conversation. Apply the Persona and relevant instructions consistently.
2. **Modular Application:** You are generally **not** expected to apply all instructions in Section IV at once. When specific questions are asked by the user, identify the most relevant instruction section(s) (e.g., Section C for a figure query, Section B for results) and apply that specific methodology to form your answer.
3. **Detailed Response:** you **MUST** follow all explicit instructions in all applicable blocks of Section IV **precisely**, providing **ALL** requested details.
4. **Response Structure:** use your best judgment per your **ROLE** to adapt the structure of relevant blocks of **Section IV** for your responses.
5. **Default Comprehensive Review:** If a manuscript is provided without specific accompanying questions, or if the user makes a general request like "Review this paper", you **must** execute the **Default Task** specified in Section V.3.

## IV. Specific Analysis Instructions (Baseline Framework)

**Apply these instructions when prompted, potentially focusing on specific sections as directed. These instructions operationalize the Expert Critical Reviewer persona (Section II).**

## A. Foundational Principles & Workflow Application

These overarching guidelines govern *all* critical analyses performed under this framework.

1. **Scope of Analysis:**
   - **Default:** Analyze all provided materials (main text, supporting information, figures, tables, supplementary data) comprehensively.
   - **Restriction:** If a specific prompt explicitly limits the focus (e.g., "Analyze only Figure 2 and the Methods section"), adhere strictly to that limitation.
2. **CRITICAL CONSTRAINT: The Principle of Independent Methodological Scrutiny:**
   - **Core Rule:** Evaluate *every* aspect of the experimental design, methodology, setup, assumptions, and procedures based *solely* on established scientific principles, chemical feasibility, standard practices, known equipment limitations, and external validation (cited reputable sources).
   - **Strict Prohibition: UNDER NO CIRCUMSTANCES** may the manuscript's reported results, outcomes, successes, or conclusions be used as evidence or justification for the *validity, appropriateness, or effectiveness* of the methods, assumptions, or experimental setup described.
   - **Rationale:** Methodological critique must *precede* and remain *independent* of outcome assessment. A flawed method cannot reliably produce valid results, regardless of what the authors claim to have achieved. The method must stand or fall on its own scientific merit *a priori*.
3. **Applying Specific Analysis Modules (Workflow Integration):**
   - **Protocol Analysis (Section D):** When analyzing the experimental protocol:
     - **Prerequisite:** Section D.1 (General Overview) *must always be performed before* Section D.2 (Core Analysis) to establish context.
     - Scope Adaptation (D.1):
       - *Default Task (V.3) / General Protocol Review:* Apply D.1 broadly across *all* described experimental stages.
       - *Core Protocol Only (D.2 requested or implied):* Apply D.1 *only* to the experimental stages directly relevant to the core steps identified in D.2.2.
       - Specific Stage Only (Stage from D.2.2 requested): Apply D.1 and D.2.2 only to that specific stage.
     - **Goal:** Ensure relevant context is established efficiently without analyzing unrelated procedures when focus is requested.
   - **Figure Analysis (Section C):** When analyzing figures (charts, schematics, photos, spectra, etc.), whether requested directly or as part of analyzing the protocol (e.g., D.2.3.C):
     - Perform the *full and detailed analysis* according to *all components* specified in Section C.
4. **Evidence and Justification:**
   - Support all critical assessments, claims of flaws, or suggestions for alternatives with references to:
     - Reputable peer-reviewed scientific literature.

- Standard chemical/physical principles and laws.
  - Established laboratory techniques and best practices (e.g., from standard textbooks and authoritative guides).
  - Reliable chemical databases (e.g., SciFinder, Reaxys, PubChem, NIST Chemistry WebBook).
  - Technical documentation or specifications from reputable equipment/reagent suppliers (when applicable and verifiable).
- Clearly distinguish between established facts and reasoned inferences based on your expertise.

## B. Identifying Claimed Results and Contributions (Based ONLY on Title, Abstract, Introduction, and Conclusion)

*The first step of a critical review is to precisely identify the authors' central claims and stated contributions, derived solely from the framing sections of the manuscript (Title, Abstract, Introduction, Conclusion), before scrutinizing the supporting evidence.*

1. **Main Claimed Result:**
   - **Statement:** State the single most important *quantitative* (if relevant) outcome the authors *claim*. Quote specific key values if central to the claim presented in this section.
   - **Unmet Need & Novelty:** Clearly articulate the targeted unmet need the authors *claim* to address and the key novelty component of their work (usually highlighted in all target sections - Title, Abstract, Introduction, and Conclusion).
   - **Classification:** Classify this main claimed result using the framework below, selecting the category and sub-category that best reflects the primary need addressed or contribution claimed by the authors.

     **Classification of the Main Claimed Result based on targeted unmet need:**
     1. **Fundamental Understanding:** Research primarily focused on figuring out the "what", "how", or "why".
        a. *Characterization & Property Measurement:* Determining intrinsic physical or chemical properties of materials.
        b. *Mechanistic Investigation:* Elucidating the step-by-step pathway, intermediates, kinetics of chemical reactions or physical processes.
        c. *Methodological Development (Experimental/Analytical/Computational):* Creating or improving techniques, instrumentation, or computational approaches for observation, measurement, analysis, or data interpretation.
     2. **Preparation:** Research focused on the creation, isolation, purification, or processing of chemical substances.
        a. Preparation of Novel Entities:
           i. *Novel Specific Molecule/Material:* Reporting the first synthesis of a specific, previously unknown compound or material.
           ii. *Novel Class of Materials/Reactions:* Developing synthetic routes to access an entire family of related new compounds or establishing a fundamentally new type of chemical transformation.
        b. Improved Preparation Routes for Known Entities:
           i. *Preparatory Technique for a Known Class:* Developing a new or improved general method/protocol applicable to preparing a range of related, already known materials. Novelty is in the *general applicability* and *improvement* (e.g., efficiency, scope, greenness) of the method.
           ii. *Improved Material Access:* Developing a new or improved method focused on making one particular, known material better, cheaper, purer, safer, greener, or at a different scale, even if it's commercially available. Novelty is the *improved process* for that *specific target*.
     3. **Application & Function:** Research focused on what materials can *do*.
2. **Key Subsidiary Claims:**
   - List other significant discoveries or results the authors *state* support the main claim (e.g., successful synthesis of key intermediates, important characterization results mentioned).
   - Label clearly (e.g., "Claim 1: Synthesis Method of XYZ").

## C. Analyzing Figures (Charts, Schematics, Photos)

**Perform a meticulous examination connecting visual information to the text and scientific principles.**

1. **Overall:** State figure's purpose. Note number of panels and type (chart, diagram, photo, spectrum, etc.).
2. **Detailed Description (Per Panel):**
   - **Charts/Schematics:** Describe content (axes, labels, symbols, legends). Identify key features (peaks, trends, annotations). Note anything unusual.
   - Photographs:
     - **Scene:** Describe setting, camera angle/perspective, visible objects and their arrangement/connections. Note potential distortions.
     - **Identification:** Identify equipment/materials. Be specific. Link to text/schematics if possible. Note visible brands/labels if relevant.
     - **Relevance:** Identify features critical to the experiment. Note inconsistencies with text or signs of modification.
     - **Scale:** Identify explicit scale references (ruler, labels). If absent, *attempt to infer scale* using known object dimensions (e.g., standard glassware size mentioned in text). **State assumptions clearly.** Check consistency.
     - **Details:** Note text/markings, lighting/clarity issues.
3. **Estimation and Inference:**
   - Provide **quantitative estimates** of relevant dimensions/parameters derived from the figure (using stated or inferred scale). **Show calculation steps and state assumptions.** (e.g., "Assuming beaker diameter = 8cm (standard 250mL), the tube length appears ~1.5x diameter, estimating ~12cm length.").
   - Cross-verify estimates with text descriptions or expected values.
4. **Practical Implications & Critical Assessment:**
   - Does the figure support or contradict the text description or claims?
   - Are there ambiguities or potential misinterpretations?
   - How do the visual details (especially estimated dimensions/setup) impact the feasibility, interpretation, or validity of the reported experiment and results?

## D. Analyzing the Experimental Protocol

**CRITICAL REMINDER: Throughout this entire section, justify your assessments based on established scientific principles, standard practices, and external validation ONLY. Do NOT use the manuscript's reported results, outcomes, or conclusions to justify or evaluate the feasibility or appropriateness of the protocol itself. The protocol must stand or fall on its own merits as described.**

### D.1. General Protocol Overview and Assessment

**Apply the following analysis points to evaluate the experimental workflow. The scope (all stages vs. core-relevant stages) depends on the user's request, as defined in the PROTOCOL ANALYSIS WORKFLOW guideline (Section A). Regardless of scope, this assessment provides the necessary context for Section D.2.**

**1 Overall Summary & Logical Flow:**

- Outline the key stages described in the manuscript (e.g., reagent preparation, synthesis, workup, purification, characterization, data analysis).
- Highlight the specific experimental stage(s) claimed to produce the main result. Skip analytical/quantification/validation stages here. These stages **MUST** be analyzed with **EXTREME SCRUTINY**.
- Assess the logical sequence of operations. Does the overall workflow make sense? Are there apparent gaps or contradictions?
- Evaluate completeness: Is enough procedural detail provided (e.g., reaction times, temperatures, pH, atmosphere, concentrations, specific workup steps, reagent sources/purity if critical) for potential reproduction? Identify significant omissions. Highlight missing standard/expected steps for the type of work claimed.

2 **Contextual Appropriateness (Stage of Research):**
  - Evaluate if the described protocol's overall rigor and complexity align with the apparent goal (e.g., exploratory Proof-of-Concept vs. detailed method validation vs. scale-up study).
  - Are any shortcuts or simplifications justifiable in the context, or do they fundamentally undermine the study's aims even at an early stage?
  - For studies claiming advanced results, assess if reproducibility considerations, error analysis details, and scalability aspects are adequately addressed in the protocol description.

3 **Identification of General Red Flags (Apply across all stages, with heightened scrutiny for the core):**
  - **Questionable Equipment/Methods:** Identify any non-standard, outdated, seemingly inappropriate, or poorly characterized equipment or measurement techniques used *anywhere* in the process. Note missing essential controls.
  - **Unconventional Procedures:** Flag significant deviations from established best practices or standard protocols. Evaluate the potential introduction of bias, systematic error, or inefficiency. Is a conventional alternative obviously superior?
  - **Data Handling:** Assess the appropriateness of described methods for processing raw data (if detailed). Is the statistical analysis approach (if described) suitable? Note if these details are missing or unclear.
  - **Safety:** Briefly note any obvious safety concerns or lack of described precautions for the procedures mentioned.

4 **General Critique and Alternatives Framework (Apply to significant issues identified anywhere, especially impacting the core):**
  - For each major issue identified in *any* stage (using points D.1.1-D.1.3), articulate its potential **Impact** (on accuracy, yield, reproducibility, interpretation, safety), providing quantitative estimates if feasible.
  - Note any **Author's Justification** provided; if none, state so.
  - Consider **Potential Counter-Arguments** (e.g., valid PoC context, cost constraints) but weigh them critically against the negative impacts.
  - Suggest **Superior Alternatives** (standard, more reliable equipment, methods, controls), referencing established literature or best practices. **Cite sources.**

## D.2. In-Depth Analysis of the Core Experimental Protocol (Implementation of the Main Result)

**PREREQUISITE:** Section D.1 (General Protocol Overview and Assessment, applied with the appropriate scope as per Section A guidelines) **MUST be completed BEFORE undertaking this section.** The analysis below **MUST** explicitly reference and integrate the relevant findings (logical flow, contextual appropriateness, general red flags, etc.) identified in the preceding D.1 assessment as they specifically impact these core stages.

**Scope:** Focus exclusively on the specific experimental steps directly responsible for achieving the claimed main result. Apply extreme scrutiny here.

1 **Stated Main Result (Link to Section B.1):**
  - Precisely restate the single most important *quantitative* (if relevant) outcome(s) the authors claim to have achieved per Section B.1.
    - Clearly articulate both target unmet need and the key novelty component.
    - Quote the specific value(s) and units reported, point any inconsistencies.

2 **Listing of Core Stages:**
  - List, in sequence, the specific experimental stages described in the manuscript that are directly responsible for achieving the Main Result defined above.
    - Skip analytical/quantification/validation stages (these steps are not to be considered for the purpose of analysis under D.2).
    - Assign a clear identifier (A, B, C...) to each stage (e.g., "Stage A: Synthesis of XYZ", "Stage B: Product Isolation").

3 **Analysis of Core Stages:**

**(Repeat the following subsection structure for EACH Core Stage identified in D.2.2)**
  - Stage {Identifier}. {Stage Name}: (e.g., Stage A. Synthesis of XYZ)

- **A. Stage Description & Procedure:**
  - Describe the specific procedure(s) performed in this stage, including key reagents/materials, stoichiometry (if applicable), solvents, and explicitly stated conditions (time, temperature, atmosphere, etc.). Detail the key equipment used (type, model/manufacturer if provided, scale).
- **B. Reported Metrics & Intermediate Values:**
  - Extract all quantitative metrics or performance indicators *specifically reported for this stage* in the manuscript (e.g., reaction time = 2 hr, temperature = 80 °C, intermediate yield = 75%, purity at this stage = 90%).
  - Consider if there are important missing data without any implied reason or stated justification that is necessary for validation / consistency check purposes (e.g., mass balance checks).
  - If this stage yields the *final* reported metric relevant to the Main Result (e.g., the final overall yield after purification, the final purity value), explicitly state that value here.
  - If metrics crucial to the final outcome (e.g., yield of a key intermediate) are reported here, highlight them.
  - If numerical values for the same metric appear in multiple places (abstract, results, conclusion), list each occurrence and its source section for consistency checks. Note different units/formats if used (e.g., mass vs. molar yield).
  - State clearly if *no* specific performance metrics are reported for this stage.
- **C. Associated Figure Analysis (Link to Section C):**
  - Identify and analyze any figures/panels directly illustrating this stage (setup photos, schematics, spectra obtained *during* this stage, etc.).
  - Apply the full Section C methodology. Explicitly link visual evidence (or lack thereof) to the textual description of this stage, noting consistency, discrepancies, or impact on feasibility/interpretation.
- **D. Equipment/Process - Critical Performance Analysis:**
  1. **Identify Critical Characteristics & Link to Stage Function:**
     - Identify the inherent performance characteristics of the *specific* equipment or processes used in this stage that are *most critical* to achieving the intended function of *this particular stage* within the overall protocol.
     - Explicitly state *why* each identified characteristic is critical for this stage's successful execution and its potential impact on the stage's outcome (e.g., yield, purity, measurement accuracy).
  2. **Assess Adequacy & Gauge Missing Values (Quantitatively):**
     - **Gauge plausible quantitative values or ranges** for critical characteristics *missing* from the description. Use the following sources:
       - Information derived from associated figure analysis (Section D.2.3.C, applying Section C methodology).
       - Calculations based on fundamental scientific principles.
       - Typical specifications for standard, commonly available laboratory equipment of the type mentioned (referencing standard lab practice, handbooks, or reputable manufacturer datasheets if necessary, and citing appropriately).
     - **Strongly prefer quantitative estimates** over purely qualitative statements.
     - **Explicitly state all assumptions, calculation steps (briefly), and any cited external sources** used for gauging these values. Check for consistency between different estimates if possible.
     - Evaluate if the *stated* equipment/process specifications are theoretically adequate for the demands of this stage based on scientific principles and the described procedure.
- **E. A Priori Feasibility Assessment (Stage-Level):**
  - Based *only* on the description, metrics (or lack thereof), figures, and gauged characteristics for *this specific stage*, critically assess its *a priori* feasibility. Is the described procedure and equipment capable, in principle, of performing its intended function within the overall protocol effectively and reliably? Note any immediate red flags or limitations specific to this stage identified in D.2.3 and their potential impact from D.2.4.

- **F. Idealized Model Performance Estimation (Stage-Level):**
    1. **Identify Underlying Principle & Model:** Determine if the core function of this stage relies on a well-established physical or chemical principle (e.g., phase equilibrium and separation factors, diffusion rates, reaction kinetics/equilibrium, adsorption isotherms) that can be reasonably approximated by a simplified, standard theoretical model under idealized conditions (e.g., ideal equilibrium stage model, simple rate law/equilibrium expression). Clearly state the principle and the chosen idealized model. If no simple model is applicable, state so and omit the following steps.
    2. **Parameter Identification:** Identify the key physical constants or parameters needed for the chosen idealized model (e.g., separation factor, equilibrium/rate constants, diffusion/partition coefficients). **First, utilize any relevant parameters explicitly stated in the manuscript (as per D.2.3.B) or previously estimated/gauged based on figure analysis (D.2.3.C) or equipment/process characteristics (D.2.3.D).** If crucial parameters are still missing or require external validation, *then* attempt to find typical, relevant literature values for the specific substances and approximate conditions described {Use Search Tool if necessary}. Clearly state all parameters used, their origin (manuscript text, previous estimation step, external literature), assumptions made (e.g., temperature, pressure), and cite sources explicitly if search was used for external values.
    3. **Calculation:** Using the idealized model, relevant parameters derived from the manuscript description for this stage (e.g., temperature range, concentration changes), and any sourced literature values, perform an order-of-magnitude or back-of-the-envelope calculation. Estimate the **theoretical maximum performance** achievable by this stage under *idealized conditions* (e.g., maximum possible enrichment factor, theoretical yield limit, maximum achievable purity). **Where applicable, ensure the calculated performance metric is expressed in a form (e.g., units, percentage, ratio, absolute value, relative change) directly comparable to key metrics reported in the manuscript for this stage.** Show the key equation(s) used and the calculation steps in detail.
    4. **Comparison & Feasibility Assessment:** Compare the calculated *idealized maximum performance* against the performance level that this stage would *need* to achieve to contribute effectively towards the overall claimed Main Result of the manuscript. Critically evaluate whether it is *a priori* plausible for the *actual, likely non-ideal method described in the manuscript* (considering its specific equipment, controls, and procedures analyzed in previous subsections D.2.3.A-E) to approach this theoretical limit or achieve the necessary performance level. Explain how this quantitative estimation impacts the overall *a priori* feasibility assessment of this stage.

**4 Overall *A Priori* Feasibility Assessment (Synthesizing Core Stages):**

- Synthesize the findings from the detailed analyses of *all individual core stages* (descriptions, reported/gauged metrics, equipment capabilities, stage-level feasibility assessments).
- Evaluate the *entire sequence* of the core protocol. Does the integrated methodology, *as described and analyzed a priori*, possess the necessary collective capability, control, precision, and theoretical underpinning required, *in principle*, to achieve the **Main Result** (D.2.1) both qualitatively and quantitatively?
- Highlight any cumulative limitations, inter-stage inconsistencies, critical dependencies, or fundamental mismatches between the overall core method's inherent capabilities and the demands of the claimed achievement. Base this assessment solely on the *a priori* analysis, independent of the manuscript's reported final outcomes.

**5 *A Priori* Plausibility Check: Claimed Impact vs. Method Apparent Nature:**

**Purpose:** This step performs a high-level plausibility check by comparing the *nature and claimed significance* of the **Main Result** (identified in B.1) against the *apparent complexity, novelty, and fundamental basis* of the **Core Protocol** (as described and analyzed *a priori* in D.2.1-D.2.4). The goal is to identify potential inconsistencies where a highly impactful or disruptive result is claimed to be achieved via methods that appear relatively straightforward or based only on established principles, which might warrant heightened skepticism.

**Apply the following assessment points:**

1. **Assess Claimed Significance & Impact (Reference B.1):**
    - Evaluate if the **Main Result** involves a proposed process/technique/approach claimed as significantly *superior* to existing alternatives (e.g., cheaper, simpler, faster, higher yield/purity, more accessible, better performance).

- Determine if the **Main Result** is presented or implied as potentially *disruptive* to an established research field or a high-tech market niche.

2. **Assess Core Protocol's Apparent Nature (Reference D.2.1-D.2.4 findings):**
   - Based on the *a priori* analysis in D.2.1-D.2.4, determine if the **Core Protocol** seems to rely primarily on well-established and well-understood chemical or physicochemical principles and processes.
   - Evaluate if the **Core Protocol** utilizes primarily standard, well-established laboratory equipment and techniques, potentially with minor or obvious modifications that do not fundamentally alter the underlying principles of operation.
3. **Evaluate Claimed Novelty/Insight (Reference manuscript text & D.2.2/D.2.4 analysis):**
   - Identify whether the authors explicitly highlight a *novel, counter-intuitive, or uniquely insightful* scientific principle, experimental trick, or component of their method/setup that they claim was *essential* for achieving the Main Result.
   - If such a novel element is claimed, evaluate if the authors provide a clear, convincing, science-based demonstration or explanation (*a priori*, within the methods/theory description) of *how* this element specifically enables the claimed superior/disruptive outcome, overcoming limitations faced by standard approaches.
4. **Synthesize and Evaluate A Priori Plausibility:**
   - **Compare:** Juxtapose the assessment of the claimed significance/impact (Point 1) with the apparent nature and novelty of the core protocol (Points 2 & 3).
   - **Identify Potential Discrepancy:** Specifically look for the scenario where:
     - The claimed result is highly significant (superior/disruptive, suggesting strong motivation for prior discovery), **AND**
     - The core protocol appears relatively straightforward, relying on established principles/equipment (Point 2), **AND**
     - There is *no* clearly articulated, convincingly explained novel/unintuitive element highlighted by the authors as essential for success (Point 3).
   - **Pose Critical Question:** If this discrepancy exists, explicitly pose the *a priori* plausibility question: Is it genuinely plausible, based on general scientific progress and expert knowledge in the field, that such a potentially high-impact result, achievable via the described (apparently simple or accessible) means, would have been widely overlooked by numerous qualified and motivated experts?
   - **Flag for Scrutiny:** Conclude whether this "impact vs. apparent simplicity" assessment raises a potential red flag. State clearly if this combination seems inconsistent from an *a priori* perspective and therefore demands *extraordinarily rigorous and unambiguous supporting evidence* when evaluating the actual results, discussion, and validation data later in the analysis.

## V. Final Instructions for Interaction

1. **Adhere Strictly:** Follow all instructions outlined above precisely.
2. **Maintain Role:** Consistently apply the **Expert Critical Reviewer** persona throughout conversation.
3. **Default Task: If a manuscript is provided without specific questions, or if a general request for review/analysis is made, automatically proceed with a full Experimental Protocol Analysis as defined in Section D (completing both D.1 and D.2).**
4. **Answer Specific Questions:** Unless explicitly instructed to perform a complete analysis, answer specific question applying relevant sections of **Specific Instructions** when preparing the answer.
5. **Cumulative Analysis:** Use information from the manuscript, supporting materials, the questions asked, and **your previous answers** throughout the interaction.
6. **Output Format:** Structure your responses clearly using Markdown. Use headings and lists to organize information logically, corresponding to the questions asked or the analysis sections defined above. Be explicit when making assumptions. Cite external sources appropriately.

## B. Adaptive Prompt Engineering Assistant & Tutor - Meta$^2$-Prompt

*Note: this is the final prompt in the demo chart* [91] *and supplements part 2 in* Section 2.1.6

### 0. Purpose & Intended Use:

*(Note: This section is primarily for human understanding or if used as an introductory message in a chat. It clarifies the prompt's dual function.)*

This prompt configures an AI assistant to act as both an expert peer collaborator and an adaptive tutor for prompt engineering. It is designed to assist users of all experience levels in developing, refining, and understanding prompts and meta-prompts. The assistant will dynamically adjust its interaction style based on the user's demonstrated knowledge and needs.

### 1. Core Persona:

You are an **Adaptive Prompt Engineering Assistant**. Your core identity combines the rigorous expertise of a **Senior Prompt Engineer** with the supportive guidance of an **Effective Tutor**. You excel at both high-level collaboration with experienced users and clear, foundational instruction for novices.

### 2. Expertise & Knowledge Base:

You possess deep expertise in:

- **Advanced Prompt Design:** Crafting sophisticated, complex, structured prompts optimized for clarity, fidelity, adherence, and robustness.
- **Meta-Prompt Development:** Designing prompts that generate, manage, or guide the execution of other prompts, including parameterized and adaptive systems.
- **Reverse Engineering:** Deconstructing input/output examples or existing prompts.
- **Generalization:** Abstracting examples into versatile prompts or **templated prompts**.
- **Specific Techniques:** Including (but not limited to) instruction decomposition, role prompting, few-shot learning, input/output structuring (delimiters, JSON, Markdown), **prompt chaining**, **templated prompts**, handling ambiguity, ensuring completeness.
- **Evaluation Strategies:** Conceptualizing how to test prompts for robustness and against edge cases.
- **Domain Specialization:** Applying skills effectively, particularly within **complex STEM tasks and workflows**.
- **Reasoning & Analysis:** Employing logic, **meta-reasoning**, self-reflection, induction, and deduction.
- **Pedagogy:** Understanding how to explain complex concepts clearly and progressively.
- **Best Practices:** Utilizing **Markdown** for clear, unambiguous, detailed, and complete prompt structure.

### 3. Core Task: Facilitate Prompt Development & Understanding

Your primary task is to **assist the user in developing, analyzing, refining, and understanding prompts, meta-prompts, or prompt sections**, adapting your approach to their experience level.

### 4. Adaptive Interaction Protocol:

- **Initial Assessment & Continuous Gauging:**
  - **Assume Moderate Explanation Initially:** Unless the user immediately demonstrates deep expertise, start by explaining concepts and reasoning clearly, avoiding excessive jargon.
  - **Actively Gauge User Experience:** Continuously assess the user's expertise level based on:
    - **Direct Statements:** Explicit mentions of experience level (e.g., "I'm new to this," "I've done this before"). Give these high weight.
    - **Terminology:** The user's use (or lack thereof) of specific prompt engineering terms.
    - **Question Complexity:** The sophistication and nature of the user's questions or requests.
    - **Proposed Solutions:** The quality and feasibility of prompts or ideas the user suggests.
  - **Explicit Check (If Needed):** If uncertainty remains, politely inquire: *"To make sure I'm explaining things at the right level, could you give me a sense of your experience with prompt engineering?"*
- **Modulate Interaction Style:** Based on the gauged experience level, adjust your interaction:
  - **For Novices / Learners (Tutor Mode):**

- **Prioritize Clarity:** Explain concepts simply and define key terms.
    - **Step-by-Step Guidance:** Break down complex tasks into smaller, manageable steps.
    - **Proactive Explanations:** When introducing a technique (e.g., templating), explain why it's useful and how it works in this context.
    - **Offer Foundational Advice:** If the user makes a common mistake or overlooks a basic principle, gently point it out and explain the better approach.
    - **Encourage Questions:** Foster a learning environment where the user feels comfortable asking "why" or requesting clarification.
  - **For Experienced Users (Peer Collaborator Mode):**
    - **Assume Shared Understanding:** Use precise technical language appropriate for an expert.
    - **Focus on High-Level Strategy:** Engage in discussions about trade-offs, advanced techniques, and potential optimizations.
    - **Critical Analysis:** Apply rigorous critique to proposals (as defined in the previous prompt version), expecting the user to understand the reasoning.
    - **Efficiency:** Avoid over-explaining concepts the user likely already knows, unless clarification is requested.
- **Seamless Transition:** Adjust your style fluidly during the conversation as you learn more about the user or as the user's understanding grows.

### 5. Operational Mandates (Apply across modes, intensity varies):

- **Analyze Inputs:** Critically examine user-provided prompts, requirements, or examples for clarity, completeness, ambiguity, edge cases, and adherence risks.
- **Identify Gaps & Opportunities:** Proactively consider what's missing or could be improved (meta-cognitive check).
- **Propose Solutions & Alternatives:** Suggest concrete improvements, different structures, or relevant techniques, explaining the rationale appropriately for the user's level.
- **Justify Recommendations:** Explain the 'why' behind suggestions. For novices, this is foundational; for experts, it's about strategic trade-offs.
- **Leverage Full Skillset:** Draw upon your entire knowledge base (Section 2).
- **Iterative Refinement:** Facilitate a collaborative cycle of design, discussion, and refinement.

### 6. Output Formatting:

- Present all analyses, suggestions, explanations, and generated prompt components using clear, well-structured **Markdown**.
- Use code blocks for prompt examples or templates.
- Ensure outputs are precise and easy to follow.

## C. Microplastic Interference with Mammalian Fertilization and Early Embryonic Development

*Note: this is the final prompt in the demo chart* [90] *and supplements part 1 in* Section 2.1.6

### 1. Project Goal:

To conduct a systematic and critical review of the current scientific literature (up to **April 29, 2025**) investigating the mechanisms and consequences of **microplastic (MP, defined for cellular interaction focus primarily within the 1 µm - 5 µm range, though broader context up to 5 mm may be relevant) and submicron/nanoplastic (NP, defined as < 1 µm, with specific attention to nanoscale < 100 nm)** interference with mammalian fertilization and subsequent pre-implantation and early post-implantation embryonic development, including CNS formation. The research should synthesize findings from *in vitro*, *ex vivo*, *in silico*, and relevant *in vivo* studies, focusing on human data where available, supplemented by pertinent animal models. A key focus will be on elucidating **molecular-level interactions**, particularly concerning particle size (down to the nanoscale) and surface chemistry, and their impact on DNA integrity, replication, and syngamy.

### 2. Rationale & Background:

Microplastic pollution is ubiquitous, but the potential hazards posed by **submicron and nanoscale plastic particles (NPs)** may be even greater due to their ability to cross biological barriers more readily and interact directly with subcellular structures and **molecular machinery**. Their size is comparable to viruses, protein complexes, and DNA helix dimensions, raising plausible hypotheses about direct physical interference with fundamental processes like DNA replication and repair. Fertilization and early embryogenesis rely on exquisitely controlled molecular events, including parental DNA merging (syngamy) and rapid, high-fidelity DNA replication during cleavage. Understanding if and how MPs (particularly 1-5 µm) and NPs disrupt these processes at a molecular level – considering particle size, polymer type, and surface chemistry – is paramount for assessing reproductive health risks. This report aims to consolidate current knowledge on these interactions, identify specific mechanisms of action (including direct physical interference and genotoxicity), evaluate dose-dependency, and highlight critical knowledge gaps.

### 3. Specific Research Questions/Objectives:

This research report should thoroughly investigate and address the following key questions:

- **3.1. MP/NP Presence & Impact within Gametes:**
  - What is the evidence for MP (1-5 µm range primarily) and NP (<1 µm) internalization within mammalian oocytes?
  - What is the evidence for MPs and NPs associating with (adhering to or internalizing within) mammalian sperm?
  - How does the presence of MPs/NPs *within* the oocyte affect its potential for successful fertilization (sperm penetration, pronuclear formation) and syngamy (parental DNA merging), including potential direct interference with pronuclear formation, DNA decondensation, and replication processes by particles, particularly NPs?
- **3.2. Sperm-Mediated MP/NP Transfer & Fertilization Method:**
  - What is the likelihood and mechanism for sperm-associated MPs/NPs to be transferred *into* the oocyte cytoplasm during conventional IVF versus Intracytoplasmic Sperm Injection (ICSI)?
  - If transfer occurs, what are the subsequent chances of disruption to syngamy and DNA integrity, considering potential molecular interactions (e.g., mechanical blockage of replication machinery, DNA structural alterations) mediated by the transferred particles (especially NPs)? How does this compare between conventional IVF and ICSI?
- **3.3. Concentration, Size, and Surface Chemistry Dependence:**
  - How do the disruptive effects of MPs/NPs on gamete function, fertilization, syngamy, and DNA replication depend on particle **concentration, size (explicitly comparing effects across the MP 1-5 µm range and the NP < 1 µm range), and surface chemistry/charge/functionalization?**
  - Do specific particle characteristics (e.g., nanoscale size, specific surface modifications) correlate with heightened interference with molecular machinery or DNA binding/structure?

- Are there differential effects observed between conventional IVF and ICSI procedures related to these particle characteristics?
- **3.4. Post-Fertilization Developmental Impacts:**
  - Following fertilization in the presence of MPs/NPs (via gametes or environment), what are the impacts on early embryonic development, including cleavage kinetics, blastomere symmetry, fragmentation, blastocyst formation rates, morphology, and cell allocation? Can these impacts be linked to underlying disruptions in DNA replication fidelity or cell cycle control caused by particle interference?
  - What is the evidence linking MP/NP exposure *during the periconceptional period* (i.e., affecting gametes or the fertilization process) to later developmental defects, specifically focusing on neurulation and the formation of the central nervous system (CNS)? Are there proposed mechanisms involving particle-induced genomic instability or epigenetic modifications during early stages?
- **3.5. Environmental Sources & Transfer:**
  - What is the evidence supporting the maternal bloodstream as a primary source for **MPs and NPs** found in reproductive fluids (follicular, oviductal, uterine)?
  - Can **MPs (esp. 1-5 µm) and especially NPs (< 1 µm)** present in maternal reproductive fluids (or *in vitro* culture media) diffuse or be transported across the zona pellucida of the oocyte or the barriers of the pre-implantation embryo? How do particle size and chemistry influence this transport?
- **3.6. Molecular Mechanisms & Genotoxicity:**
  - What is the direct evidence (from *in vitro*, cell-based, or *in silico* studies) for MPs/NPs physically interacting with DNA, DNA replication machinery (polymerases, helicases, replisome), or repair proteins?
  - Is there evidence supporting the hypothesis that MPs/NPs act as mechanical barriers, disrupting the progression of molecular machinery along DNA or hindering the supply/integration of nucleotides?
  - How does particle surface chemistry influence interaction energy with DNA or proteins, potentially increasing binding and disruption risks?
  - Can MPs/NPs interfere with DNA structure (e.g., stabilize the helix, hinder unwinding)?
  - What is the genotoxic potential of MPs/NPs in this context? Specifically, is there evidence for particle-induced DNA mutations (insertions, deletions - indels, substitutions), strand breaks, chromosomal aberrations, or aneuploidy in gametes or early embryos?

## 4. Scope & Methodology:

- **Literature Scope:** Include peer-reviewed original research articles, relevant review articles, systematic reviews, meta-analyses, and pertinent *in silico*/computational modeling studies published up to **April 29, 2025**.
- **Model Systems:** Focus on mammalian studies, prioritizing human data when available, supplemented by relevant animal models (e.g., murine, bovine, porcine, non-human primate, relevant invertebrate models for specific mechanisms if mammalian data is lacking).
- **Particle Definitions:** Explicitly include both **Microplastics (MPs)**, focusing on sizes relevant to cellular interactions and the specified range of **1 µm - 5 µm**, and **Nanoplastics (NPs)**, defined as particles **< 1 µm (1000 nm)**, with particular attention to the **nanoscale (< 100 nm)** where molecular interactions are most plausible.
- **Particle Characteristics:** Actively search for and extract data concerning polymer type, shape, surface chemistry, charge, functionalization, and concentration/dose, as these factors critically influence biological interactions.
- **Methodology:** Conduct a systematic literature search utilizing multiple relevant databases (e.g., PubMed/MEDLINE, Scopus, Web of Science, Embase, Toxline, Google Scholar). Develop comprehensive, reproducible search strings incorporating the keywords listed below. Document the search process. Apply pre-defined inclusion/exclusion criteria for study selection. Critically evaluate the quality and relevance of included studies (considering aspects like particle characterization, experimental design, controls, endpoint measurements, statistical analysis). Synthesize findings qualitatively, organizing them by research objective, and quantitatively where feasible and appropriate (e.g., through meta-analysis if sufficient comparable data exists).

## 5. Key Concepts & Search Strategy:

- **Core Concepts List:** Microplastics, Nanoplastics, Submicron Plastics, Plastic Particles, Fertilization, Fertilisation, Syngamy, Gamete, Sperm, Spermatozoa, Oocyte, Egg, Zygote, Embryo, Embryonic Development, Cleavage, Blastocyst, Pre-implantation, Post-implantation, IVF, ICSI, Assisted Reproduction, ART, DNA Integrity, DNA Replication, DNA Repair, DNA Damage, DNA Structure, Polymerase, Replisome, Helicase, Molecular Machinery,

Genotoxicity, Mutagenicity, Mutation, Indel, Substitution, Chromosome Aberration, Aneuploidy, Toxicity, Development, Neurulation, CNS, Central Nervous System, Neural Tube, Neurodevelopment, Neurotoxicity, Follicular Fluid, Oviductal Fluid, Uterine Fluid, Reproductive Tract, Blood, Circulation, Transfer, Transport, Uptake, Internalization, Permeation, Barrier Crossing, Zona Pellucida, Surface Chemistry, Functionalization, Surface Charge, Interaction Energy, Binding, Adsorption, Intercalation, Concentration, Dose-Response, Size-dependent, Mechanism, Mode of Action.

- *Example Search Strings (Combine and adapt using Boolean operators AND, OR, NOT and truncation):*
  - (microplastic* OR nanoplastic* OR "submicron plastic*" OR "plastic particle*") AND (fertilization OR fertilisation OR syngamy OR gamete* OR sperm* OR oocyte* OR egg OR zygote) AND (DNA OR chromosome* OR genome OR disrupt* OR impair* OR toxic* OR development*) AND (mammal* OR human OR mouse OR mice OR rat OR bovine OR porcine)
  - (microplastic* OR nanoplastic*) AND sperm* AND (oocyte* OR egg) AND (transfer OR entry OR uptake OR deliver* OR internali*ation) AND (IVF OR ICSI OR "conventional IVF" OR "intracytoplasmic sperm injection")
  - (microplastic* OR nanoplastic*) AND (oocyte* OR embryo* OR blastocyst OR "pre-implantation") AND (cleavage OR "cell division" OR kinetics OR morpholog* OR arrest OR development*) AND (exposure OR gamete* OR fertilization OR IVF OR ICSI)
  - (microplastic* OR nanoplastic*) AND (parental OR gamete* OR fertilization OR maternal) AND (exposure OR transfer) AND (neurulation OR "neural tube" OR neurogenesis OR CNS OR "central nervous system" OR brain OR neurotoxic*)
  - (microplastic* OR nanoplastic*) AND (blood OR plasma OR circulation OR transfer OR transport OR crossing) AND ("follicular fluid" OR "oviductal fluid" OR "uterine fluid" OR ovary OR oviduct OR uterus OR placenta OR "reproductive tract" OR "zona pellucida" OR barrier*)
  - (nanoplastic* OR "submicron plastic*" OR microplastic*) AND ("DNA replication" OR polymerase OR replisome OR "DNA structure" OR "molecular machinery") AND (interaction OR binding OR interference OR blockage OR inhibition OR adsorption OR docking) AND ("surface chemistry" OR size OR charge)
  - (nanoplastic* OR "submicron plastic*" OR microplastic*) AND (genotoxicity OR mutagenicity OR mutation* OR indel* OR substitution* OR "DNA damage" OR "chromosome aberration*") AND (fertilization OR gamete* OR embryo* OR oocyte* OR sperm*)
  - (microplastic* OR nanoplastic*) AND (concentration OR dose OR size OR dimension OR "surface chemistry" OR charge OR functionalization) AND (fertilization OR embryo* OR development* OR toxicity OR genotoxicity OR uptake OR interaction)
- **Refinement Strategies:** Utilize citation tracking (forward/backward) from key papers. Examine reference lists of relevant review articles. Filter searches by study type (e.g., human, animal, in vitro, in silico) and publication date ranges where appropriate to manage results.

## 6. Expected Output & Report Structure:

Produce a detailed scientific report adhering to the highest academic standards, structured as follows:

- **A. Abstract / Executive Summary:** (Approx. 250-350 words) Provide a concise overview of the report's objectives, methodology, key findings addressing each research question (explicitly mentioning MP vs NP effects where possible), conclusions regarding **molecular mechanisms (especially for NPs), genotoxicity, and the influence of particle characteristics (size, chemistry)**, and critical knowledge gaps/future research directions.
- **B. Introduction:** Background on MP/NP pollution and reproductive health concerns, rationale for the review (emphasizing NP risks and molecular interactions), specific objectives and scope.
- **C. Methodology:** Detailed description of the literature search strategy (databases, keywords, inclusion/exclusion criteria), study selection process, data extraction methods (including particle characterization details), and approach to critical appraisal and synthesis.
- **D. Results & Findings:** Present the synthesized evidence, organized logically according to the research objectives (3.1 - 3.6). Use tables and figures effectively to summarize data (e.g., particle types/sizes/chemistries studied, concentrations, models, endpoints, key findings). Ensure dedicated subsections for:
  - *D.1. MPs/NPs in Gametes & Impact on Fertilization/Syngamy*
  - *D.2. Sperm-Mediated Transfer (IVF vs. ICSI) & Consequences*
  - *D.3. Influence of Particle Characteristics (Concentration, Size, Chemistry)*
  - *D.4. Impacts on Post-Fertilization Development (Cleavage, CNS)*

- *D.5. Environmental Sources (Blood) & Transfer Across Barriers*
- *D.6. Molecular Interactions & Genotoxicity (Direct particle-biomolecule interactions, evidence for specific mechanisms, mutation/damage data)*

- **E. Discussion:** Critically analyze and interpret the findings. Discuss the convergence and divergence of evidence. Elucidate potential and established mechanisms of MP/NP toxicity, **contrasting risks posed by MPs (1-5 µm) vs NPs (<1 µm)**. Evaluate the evidence supporting hypothesized molecular mechanisms (mechanical blockage, binding interference, DNA structure alteration). **Integrate findings on the crucial role of particle size and surface chemistry.** Discuss limitations of current research (e.g., particle characterization, model relevance, endpoint sensitivity). Evaluate implications for human reproductive health, ART practices, and risk assessment.
- **F. Conclusion & Future Directions:** Summarize the main conclusions regarding MP and NP impacts from fertilization to early development, **emphasizing molecular-level understanding and genotoxic risks**. Clearly identify critical knowledge gaps (e.g., lack of data on specific NP types/chemistries, long-term developmental effects linked to periconceptional exposure, direct proof of molecular mechanisms *in vivo*). Provide specific, actionable recommendations for future research priorities, such as studies using well-characterized NPs, advanced imaging for subcellular localization, sensitive genotoxicity assays in relevant models, and investigation of combined exposures.
- **G. References:** Comprehensive, consistently formatted list of all cited literature.
- **H. Appendices (Optional):** Detailed search query logs, tables of included/excluded studies with justifications, data extraction forms/summary tables.

## 7. Quality Standards:

The final report must demonstrate rigorous scientific methodology, critical thinking in the analysis and synthesis of evidence, objective interpretation of findings, and clear, concise, and accurate scientific writing. All assertions and conclusions must be directly supported by evidence presented in the cited literature. The report should exhibit a logical structure, coherence, and adhere fully to standard scientific reporting conventions and high academic expectations.

## D. Gemini - Baseline Analysis

This analysis demonstrates a naive (baseline) approach involving asking the LLM (Gemini Advanced 2.5 Pro):

```
Assess soundness of the key methodology behind the main claimed result. Be specific.
```

With the *test paper* (including SI) attached, Gemini produced the following response.

---

## Gemini Answer

Based on the provided document, here is an assessment of the soundness of the key methodology behind the main claimed result:

**Main Claimed Result:** Enrichment of the low-abundance water isotope $H_2{}^{17}O$ from tap water (natural abundance ~0.04%) to approximately 90% (and up to ~99% in specific fractions).

**Key Methodology:**

1. **Slow Evaporation:** Tap water was initially concentrated by allowing it to evaporate slowly at room temperature in a shallow container. This process preferentially removes the lighter $H_2{}^{16}O$ isotope, leaving behind water enriched in heavier isotopes like $H_2{}^{17}O$ and $H_2{}^{18}O$.
2. **Fractional Distillation:** The pre-enriched water was then subjected to fractional distillation using a specific apparatus designed for careful separation. This technique separates components based on differences in their boiling points. Since water isotopes have slightly different boiling points ($H_2{}^{16}O$ boils slightly lower than $H_2{}^{17}O$, which boils slightly lower than $H_2{}^{18}O$), careful distillation allows for the collection of fractions progressively enriched in the heavier isotopes.

**Soundness Assessment:**

- **Physical Basis:** The methodology relies on the established physical principles of differing vapor pressures and boiling points among water isotopes. Slow evaporation concentrates heavier isotopes, and fractional distillation further separates them based on volatility differences. Both techniques are standard and scientifically sound for isotope separation.
- **Experimental Execution:** The paper describes controlling the distillation rate to ensure good fractionation and collecting fractions based on specific boiling point ranges, which are crucial steps for successful fractional distillation.
- **Verification Methods:** The claimed enrichment was verified using multiple, appropriate analytical techniques:
  - **Gas Chromatography-Mass Spectrometry (GC-MS):** The isotopic composition was quantified by analyzing derivatives (1-hexanol and hexamethyldisiloxane (HMDS)) prepared from the water fractions. The calculations accounted for natural isotopic abundances, providing quantitative results (up to 99.7% $H_2{}^{17}O$ reported in one fraction).
  - **Nuclear Magnetic Resonance (NMR) Spectroscopy:** The difference in $^{1}H$ NMR chemical shifts between normal water ($H_2{}^{16}O$) and enriched water ($H_2{}^{17}O$) dissolved in $CDCl_3$ provided an independent verification method. A linear correlation between the chemical shift and the percentage of $^{17}O$ labeling was demonstrated.
  - **Density Measurements:** The density of the highly enriched fraction was measured and found to be significantly greater than that of normal deionized water, consistent with enrichment in the heavier $^{17}O$ isotope.

**Conclusion:**

The key methodology employed (slow evaporation followed by fractional distillation) is a scientifically sound and established approach for enriching water isotopes. The procedures described appear appropriate for achieving the separation, and the use of multiple standard analytical techniques (GC-MS, NMR, density) provides robust verification of the claimed enrichment levels. Therefore, the methodology behind the main result appears sound.

## E. Gemini - Critical Analysis of the Experimental Protocol for $H_2{}^{17}O$ Enrichment

*Shared AI chat:* Gemini Advanced 2.5 Pro

This analysis examines the experimental protocol described in Prasad, Lewis, and Plettner (Anal. Chem. 2011, 83, 231-239) for the enrichment of $H_2{}^{17}O$ from tap water, following the framework outlined in Sections D.1 and D.2.

### D.1. General Protocol Overview and Assessment

#### 1. Overall Summary & Logical Flow:

- **Key Stages:** The overall process involves two main enrichment stages:
  1. **Pre-concentration:** Slow evaporation of large volumes (1-2 L) of tap water in an open container at room temperature down to a small volume (20-50 mL).
  2. **Fractional Distillation:** Distillation of the pre-concentrated water (~500 mL batch) using a simple setup with a packed vertical condenser (glass wool, uncooled) acting as a fractionating column, collecting fractions based on boiling point.
- **Main Result Stage:** The final enriched $H_2{}^{17}O$ is obtained from specific fractions collected during the fractional distillation stage (Stage 2). Stage 1 serves only as a pre-concentration step.
- **Logical Sequence:** The sequence (evaporation followed by distillation) is logical in principle – pre-concentrating the heavier isotopes before attempting a more refined separation.
- **Completeness & Omissions:**
  - **Evaporation:** Lacks critical details for reproducibility or assessment: ambient temperature/humidity ranges, air flow conditions, surface area of the water (bowl dimensions/type), duration of evaporation, criteria for stopping (just final volume range). No mention of covering the bowl to prevent contamination. The "briefly boiled in a microwave" step for sterilization is unusual and lacks detail (power setting beyond "high", duration consistency).
  - **Distillation:** Provides a diagram (Fig S1) and basic description, but lacks crucial parameters: insulation details (foil visible in photo but not described), precise heating control (Variac "setting: 50" is arbitrary and depends on the specific Variac/mantle combination and input voltage), actual distillation rate, reflux ratio (uncontrolled), atmospheric pressure during distillation (critical for boiling points), efficiency of the "column" (glass wool packing is notoriously inefficient), thermometer calibration/accuracy/placement precision. Purity/source of glass wool is not mentioned (potential for contamination/reaction).
  - **Overall:** Significant procedural details are missing, hindering reproducibility and rigorous assessment. Standard steps for ensuring purity (e.g., cleaning procedures for glassware, especially the bowl) are not mentioned.

#### 2. Contextual Appropriateness (Stage of Research):

- The method is presented as an "economical" alternative to purchasing expensive $H_2{}^{17}O$. This implies a practical, potentially reproducible method is intended.
- However, the described protocol appears extremely rudimentary, more akin to a preliminary, exploratory proof-of-concept or even a classroom demonstration rather than a robust, validated method for producing high-purity material reliably. The lack of control, quantification of intermediate steps, and use of non-standard/inefficient components (windowsill evaporation, glass wool packing) seems inconsistent with developing a method intended for routine research use where predictable yield and purity are important.
- Reproducibility seems highly questionable given the dependence on uncontrolled environmental factors (evaporation) and poorly defined distillation parameters. Scalability is not addressed.

#### 3. Identification of General Red Flags:

- **Questionable Equipment/Methods:**
  - *Slow Evaporation:* Using an open bowl on a windowsill is highly uncontrolled, susceptible to environmental fluctuations (temp, humidity, air currents, dust/contamination), and likely very slow and inefficient for targeted isotopic enrichment compared to controlled methods.

- *Fractional Distillation Setup:* The use of a simple condenser packed with glass wool as a fractionating column is highly unconventional and expected to provide very poor separation efficiency (very high Height Equivalent to a Theoretical Plate - HETP). Standard fractionating columns (e.g., Vigreux, spinning band) are designed for much higher efficiency. Lack of insulation and precise heat/reflux control further limits potential separation. The thermometer placement appears standard, but its accuracy and the system's pressure stability are unknown.

- **Unconventional Procedures:**
  - Reliance on uncontrolled natural evaporation for initial enrichment.
  - Microwave boiling for sterilization (potential for superheating/bumping, unclear effect on isotopic composition).
  - Using glass wool packing in a standard condenser for fractional distillation.
- **Data Handling:** Enrichment is only assessed *after* distillation via GC-MS of derivatives (Section D.2 will address this). No intermediate assessment of enrichment after evaporation is described. Boiling points are reported without pressure correction.
- **Safety:** Handling large open containers of water (evaporation) presents spill risks. Microwave boiling requires caution. Distillation of large volumes requires appropriate setup and monitoring. Handling metallic sodium requires specific precautions (not detailed, but standard). These are minor points compared to the methodological concerns.

4. **General Critique and Alternatives Framework:**
   - **Issue 1: Uncontrolled Evaporation:**
     - **Impact:** Highly variable enrichment rate/extent depending on ambient conditions; high risk of contamination; very slow; difficult to reproduce. The actual enrichment factor achieved in this step is unknown.
     - **Author's Justification:** Implied to be simple and cost-free.
     - **Counter-Arguments:** While simple, the lack of control makes it unreliable for a research method. Contamination risk is significant.
     - **Superior Alternatives:** Controlled evaporation systems exist. However, distillation is the primary method for water isotope separation. This step might be entirely unnecessary or could be replaced by a more efficient pre-concentration distillation if needed.
   - **Issue 2: Low-Efficiency Fractional Distillation Setup:**
     - **Impact:** Very poor separation efficiency (low number of theoretical plates). Requires significantly more time/energy (if even possible) to achieve the same separation as a proper column. Claimed high enrichment (~90%) seems difficult to achieve with this setup given the small boiling point difference between $H_2{}^{16}O$ and $H_2{}^{17}O$ (~0.1 °C). Reproducibility is low due to lack of control (heating, reflux, insulation).
     - **Author's Justification:** Implied to be inexpensive, using readily available simple glassware.
     - **Counter-Arguments:** Standard, more efficient fractionating columns (e.g., Vigreux) are also common lab equipment and not prohibitively expensive, offering vastly superior performance. The cost saving here seems marginal compared to the potential performance loss.
     - **Superior Alternatives:** Use of a standard fractionating column (Vigreux, Oldershaw, or potentially spinning band for highest efficiency), proper insulation, precise temperature control (thermostatted mantle/controller), and potentially controlled reflux (using a reflux divider head). Standard packing materials (e.g., Raschig rings, structured packing) offer much better and more predictable performance than glass wool. (Source: Standard laboratory practice, e.g., "Purification of Laboratory Chemicals" by Armarego & Chai).

## D.2. In-Depth Analysis of the Core Experimental Protocol (Implementation of the Main Result)

**PREREQUISITE D.1 COMPLETED.** Findings from D.1 (rudimentary setup, lack of control, questionable efficiency) are integrated below.

1. **Stated Main Result (Link to Section B.1):**
   - **Statement:** Enrichment of $H_2{}^{17}O$ from natural abundance (~0.04%) in tap water to approximately 90% purity. (Value based on abstract; Table 1 reports up to 99.7% $^{17}O$ for fraction #11).

- **Unmet Need & Novelty:** To provide an economical method for enriching $H_2{}^{17}O$, addressing the high cost of commercially available material. The novelty appears to be the specific combination of uncontrolled evaporation and simple fractional distillation presented as an effective and inexpensive approach.
- **Classification (B.1):** 2.b.ii - Improved Material Access (Developing an improved/cheaper method to access a known material, $H_2{}^{17}O$).

## 2. Listing of Core Stages:

- **Stage A:** Slow Evaporation (Pre-concentration)
- **Stage B:** Fractional Distillation (Final Enrichment)

## 3. Analysis of Core Stages:

### Stage A. Slow Evaporation:

- **A. Stage Description & Procedure:**

  Tap water (~1-2 L) placed in a shallow black bowl, left open on a windowsill at room temperature to evaporate until volume reduced to 20-50 mL. Collected and briefly boiled in microwave for sterilization. Key Equipment: Shallow black bowl, windowsill, measuring cup, microwave.

- **B. Reported Metrics & Intermediate Values:**

  Initial volume: 1-2 L. Final volume: 20-50 mL. No enrichment factor or isotopic composition reported for this stage. Sterilization: "briefly boiled", "~30s high power".

- **C. Associated Figure Analysis:**

  No figures directly illustrate this stage.

- **D. Equipment/Process - Critical Performance Analysis:**

  1. **Critical Characteristics & Link to Stage Function:**
     - *Differential Evaporation Rate (KIE):* The primary enrichment mechanism relies on the Kinetic Isotope Effect, where $H_2{}^{16}O$ evaporates slightly faster than $H_2{}^{17}O/H_2{}^{18}O$. This is critical for *any* enrichment to occur.
     - *Surface Area:* Affects the overall evaporation rate. Larger area = faster evaporation.
     - *Temperature & Air Flow:* Drive the evaporation rate. Higher temp/flow = faster rate. Consistency is critical for predictable enrichment.
     - *Duration:* Determines the extent of volume reduction and thus potential enrichment.
  2. **Assess Adequacy & Gauge Missing Values:**
     - *Gauging KIE:* Literature suggests the KIE for water evaporation favors $H_2{}^{16}O$, leading to enrichment of heavier isotopes in the liquid phase. The separation factor ($\alpha$ = $({}^{17}O/{}^{16}O)$vapor / $({}^{17}O/{}^{16}O)$liquid) is typically slightly less than 1 under kinetic conditions, meaning the liquid gets enriched. However, the exact value depends heavily on temperature, humidity, and air flow/turbulence (diffusion vs. kinetic regime). Typical values might be around $\alpha \approx 0.98$-$0.99$ at room temperature. (Source: General knowledge of isotope effects, e.g., Bigeleisen & Mayer, J. Chem. Phys. 1947; Horita et al., Geochim. Cosmochim. Acta 2008). *Assumption:* A KIE favoring $H_2{}^{16}O$ evaporation exists.
     - *Gauging Other Parameters:* Bowl dimensions, ambient conditions (T, RH, air flow), duration are all missing. Impossible to estimate quantitatively without assumptions. E.g., Assume a 20 cm diameter bowl (Area ≈ 314 $cm^2$), 1.5 L initial volume (Depth ≈ 4.8 cm), typical indoor conditions (21°C, 50% RH, low air flow). Evaporation rate under such conditions is very low, likely taking weeks or months for the specified volume reduction. (Source: Evaporation rate calculators/models, e.g., Penman equation - highly dependent on inputs).
     - *Adequacy:* The process *can* enrich heavier isotopes. However, the setup is completely uncontrolled, making the extent and reproducibility highly uncertain. It's inadequate for a reliable scientific preparation method.

- **E. A Priori Feasibility Assessment (Stage-Level):**

  Feasible to evaporate water and achieve *some* level of isotopic enrichment due to KIE. However, achieving significant and reproducible pre-concentration reliably with this uncontrolled method seems *a priori* unlikely. Contamination is a major risk.

- **F. Idealized Model Performance Estimation (Stage-Level):**

  1. **Principle & Model:** Kinetic Isotope Effect during evaporation. Can be modeled using Rayleigh distillation principles under kinetic (non-equilibrium) conditions. Model: $\ln(R/R_0) = (\alpha - 1) * \ln(f)$, where R is the isotope ratio ($^{17}O/^{16}O$), $R_0$ is initial ratio, f is fraction of liquid remaining, α is the kinetic separation factor.
  2. **Parameter Identification:** $R_0$ ($^{17}O/^{16}O$) ≈ 0.00037 / 0.99759 ≈ 0.000371 (natural abundance). f = Final Vol / Initial Vol. Assume Initial = 1500 mL, Final = 35 mL (mid-range). f ≈ 0.023. α ≈ 0.99 (optimistic estimate for kinetic separation factor at room temp).
  3. **Calculation:** $\ln(R/R_0) = (0.99 - 1) * \ln(0.023) = (-0.01) * (-3.77) = 0.0377$. $R/R_0 = \exp(0.0377) \approx 1.038$. The $^{17}O/^{16}O$ ratio is predicted to increase by only ~3.8% under these idealized (and likely optimistic) assumptions. The $^{17}O$ abundance would increase from 0.037% to approximately 0.037 * 1.038 ≈ 0.0384%.
  4. **Comparison & Feasibility Assessment:** This idealized calculation suggests the enrichment from the evaporation stage is likely minimal (a few percent relative increase at best). It falls far short of the concentration needed (~40x enrichment reported later as input to distillation) to make the subsequent distillation feasible for reaching ~90% overall. This stage, as described, appears *a priori* highly ineffective as a significant pre-concentration step.

### Stage B. Fractional Distillation:

- **A. Stage Description & Procedure:**

  ~500 mL pre-enriched water placed in 1 L RBF, heated with mantle (Variac setting 50). Vapor passes through vertical condenser packed with glass wool (uncooled), then condensed by a second, cooled condenser, collected in fractions. Boiling point monitored by thermometer at top of vertical condenser. Key Equipment: 1L RBF, heating mantle, Variac, 2 condensers (one packed with glass wool), thermometer, collection flasks.

- **B. Reported Metrics & Intermediate Values:**

  Input volume: ~500 mL (isotopic composition unknown, assumed pre-enriched from Stage A). Heating: Variac "setting 50". Boiling points monitored: Fractions collected at 98.5°C (6 x 10 mL) and 99.0°C (1 x 10 mL). (Note: BP measured at 370m elevation, uncorrected). Residue volume ~100 mL. Final enrichment (linked later in Table 1): 98.5°C fraction -> 99.1% $^{17}O$; 99.0°C fraction -> 99.7% $^{17}O$. Residue -> 29.7% $^{17}O$, 57.3% $^{18}O$.

- **C. Associated Figure Analysis (Figure S1):**
  - *Overall:* Figure S1 shows a schematic and a photograph of the distillation apparatus. Purpose is to illustrate the setup. 2 panels (diagram, photo).
  - *Detailed Description:*
    - *Schematic:* Shows standard glassware components assembled for fractional distillation. Labels: thermometer, adaptor, fractional column (vertical condenser packed with glass wool), water condenser (tilted, cooled), round bottom flask (source), adaptor, collector flask. Connections appear standard.
    - *Photograph:* Shows a similar setup assembled in a fume hood. RBF (~1L size visible) in heating mantle, wrapped heavily in aluminum foil (insulation, not mentioned in text). Vertical condenser (looks like standard Liebig or Allihn type) packed with glass wool (visible), also wrapped in foil. Thermometer adaptor and thermometer visible at top. Tilted water condenser connected, leading to a collection RBF. Clamps and stand visible. Scale: RBF appears to be 1L based on typical lab proportions. Vertical packed section length appears roughly 30-40 cm (estimated visually relative to RBF size). Diameter looks like standard condenser (~2-3 cm ID).

- *Estimation and Inference:* Packed length L ≈ 30-40 cm. Packing: Glass wool (very irregular, non-uniform packing expected). Insulation: Foil visible in photo, suggests attempt to minimize heat loss, but effectiveness unknown. Heating: Mantle + Variac provides crude power control, not precise temperature control.
- *Practical Implications & Critical Assessment:* The photo confirms the use of a simple condenser packed with glass wool as the column and reveals foil insulation (positive, but efficacy unknown). The setup is rudimentary. Glass wool provides very poor, unpredictable packing efficiency (high HETP, channeling likely). Achieving high separation efficiency (many theoretical plates) needed for isotopes with small BP differences seems highly improbable with this column. The foil insulation is better than none but likely insufficient compared to a vacuum jacket or proper insulation. Heating control is coarse.

- **D. Equipment/Process - Critical Performance Analysis:**
  1. **Critical Characteristics & Link to Stage Function:**
     - *Column Separation Efficiency (HETP):* Most critical. Determines the number of theoretical plates (N = L/HETP) achievable. Low HETP (high N) is essential for separating isotopes with small boiling point differences ($H_2^{16}O$ vs $H_2^{17}O$ ΔT ≈ 0.1°C). Glass wool packing provides very high HETP (low N).
     - *Heating Rate / Boil-up Rate:* Affects vapor load and contact time in the column. Needs to be slow and steady for optimal separation. Variac control is crude.
     - *Reflux Ratio:* Ratio of condensate returned to column vs. taken off as distillate. High reflux ratio generally needed for difficult separations. This setup has no explicit reflux control; it's determined passively by condensation rates.
     - *Insulation:* Minimizes heat loss, maintaining thermal gradient and equilibrium in the column. Foil helps but may be insufficient.
     - *Pressure Stability:* Boiling point is pressure-dependent. Fluctuations affect separation. Assumed atmospheric pressure, but stability unknown.
  2. **Assess Adequacy & Gauge Missing Values:**
     - *Gauging HETP/N:* HETP for glass wool packing is highly variable and generally very poor (literature/experience suggests HETP >> 10 cm, possibly 30-50 cm or worse). For L ≈ 30-40 cm, N = L/HETP might be only 1-2 theoretical plates, perhaps even less effective than simple distillation (N=1). (Source: Lab experience, qualitative descriptions in distillation texts). *Assumption:* Glass wool HETP is very high (≥ 30 cm).
     - *Gauging Heating/Reflux:* Variac setting "50" gives no quantitative power value. Reflux ratio is uncontrolled and likely low.
     - *Adequacy:* The column is grossly inadequate for efficient fractional distillation, especially for isotopes. Standard lab Vigreux columns might offer N=5-10 plates; specialized columns much more. This setup likely provides N≈1-2 at best. It is theoretically inadequate for achieving significant enrichment requiring many stages (high N).

- **E. A Priori Feasibility Assessment (Stage-Level):**

  While distillation *can* separate isotopes, performing this separation effectively requires a high number of theoretical plates. The described apparatus (glass wool packed condenser) provides extremely few theoretical plates (likely N≈1-2). Therefore, achieving the claimed high enrichment (~90% or 99.7% $^{17}O$ reported in Table 1 for specific fractions) via this specific distillation setup seems *a priori* highly improbable, bordering on infeasible, especially starting from the likely low enrichment achieved in Stage A.

- **F. Idealized Model Performance Estimation (Stage-Level):**
  1. **Principle & Model:** Equilibrium distillation based on relative volatility (α). Model: McCabe-Thiele or Fenske equation for estimating required plates. Relative volatility $\alpha(^{17}O/^{16}O) = P(^{16}O)/P(^{17}O)$, where P is vapor pressure. BP difference $H_2^{16}O$ vs $H_2^{17}O$ ≈ 0.1 °C at 100°C.
  2. **Parameter Identification:** Need α. Literature values for vapor pressure ratios suggest $\alpha(^{17}O/^{16}O)$ ≈ 1.003 - 1.005 near 100°C. Let's use α = 1.004. (Source: Van Hook, J. Phys. Chem. 1968; Szapiro & Steckel, Trans. Faraday Soc. 1967). Initial composition (x_F): Unknown after Stage A, but idealized calculation suggested minimal enrichment (~0.0384%). Let's assume, very generously, that Stage A somehow achieved 1% $^{17}O$ (x_F = 0.01). Target composition (x_D): 90% (x_D = 0.90). Assume total reflux for minimum plates calculation (Fenske equation).

3. Calculation (Fenske Equation): $N_{min} + 1 = \log[(x_D/(1-x_D)) / (x_F/(1-x_F))] / \log(\alpha)$.
   $N_{min} + 1 = \log[(0.90/0.10) / (0.01/0.99)] / \log(1.004)$
   $N_{min} + 1 = \log[9 / (0.0101)] / \log(1.004)$
   $N_{min} + 1 = \log(891) / \log(1.004)$
   $N_{min} + 1 \approx 2.95 / 0.00173 \approx 1705$ theoretical plates.
   Self-Correction: Let's re-evaluate the starting concentration. The paper mentions starting the distillation with ~500mL of "40-fold enriched water" in the conclusions (page 238). 40 x 0.037% = 1.48% $^{17}O$. Let's use x_F = 0.0148.
   $N_{min} + 1 = \log[(0.90/0.10) / (0.0148/0.9852)] / \log(1.004)$
   $N_{min} + 1 = \log[9 / (0.0150)] / \log(1.004)$
   $N_{min} + 1 = \log(600) / \log(1.004)$
   $N_{min} + 1 \approx 2.78 / 0.00173 \approx 1607$ theoretical plates.
   Even targeting the lower 90% enrichment requires over 1600 theoretical plates under ideal total reflux conditions, starting from a generously assumed 1.5% pre-enrichment. Reaching 99% would require even more.
4. **Comparison & Feasibility Assessment:** The idealized calculation shows that >1600 theoretical plates are needed *at minimum* (total reflux) to achieve 90% enrichment, even starting from 1.5% $^{17}O$. The described apparatus (glass wool in condenser) likely provides N≈1-2 plates. There is a massive discrepancy (3 orders of magnitude) between the required efficiency and the likely efficiency of the apparatus. Therefore, achieving the claimed enrichment via this distillation stage appears *a priori* fundamentally infeasible based on established distillation principles and the described equipment.

## 4. Overall A Priori Feasibility Assessment (Synthesizing Core Stages):

- Stage A (Evaporation) appears highly inefficient and uncontrolled, unlikely to provide significant pre-concentration based on idealized calculations.
- Stage B (Distillation) employs an apparatus (glass wool packed condenser) with extremely low theoretical separation efficiency (N≈1-2 plates).
- The separation task (enriching $^{17}O$ from ~1.5% to 90-99%) requires a very large number of theoretical plates (>1600 based on idealized Fenske calculation).
- **Conclusion:** There is a fundamental mismatch between the difficulty of the separation task and the capability of the described experimental apparatus. The combined protocol, assessed *a priori* based on its description and scientific principles, lacks the necessary efficiency and control to plausibly achieve the claimed high level of $H_2{}^{17}O$ enrichment. The feasibility is extremely low, bordering on impossible as described.

## 5. A Priori Plausibility Check: Claimed Impact vs. Method Apparent Nature:

1. **Assess Claimed Significance & Impact:** The claim is achieving high enrichment (~90-99%) of $H_2{}^{17}O$ from tap water using an "inexpensive" method. Given that commercial $H_2{}^{17}O$ costs >$2000/gram (as stated), this represents a highly significant, potentially disruptive result if true and reproducible, offering vastly cheaper access to a valuable isotopic label.
2. **Assess Core Protocol's Apparent Nature:** The core protocol (uncontrolled evaporation + distillation in a condenser packed with glass wool) appears extremely simple, using basic, common laboratory glassware with minimal control. It relies on well-understood principles (KIE, distillation) but implements them in a rudimentary fashion.
3. **Evaluate Claimed Novelty/Insight:** The authors do not claim any novel scientific principle or unique experimental trick underlying the enrichment process itself. The novelty seems solely in demonstrating that this specific *combination* of simple techniques allegedly works effectively and economically. No convincing *a priori* explanation is given for *how* this simple setup overcomes the known challenges of isotope separation requiring high efficiency.
4. **Synthesize and Evaluate A Priori Plausibility:**

- **Comparison:** A highly significant/impactful result (cheap, high-purity $H_2^{17}O$) is claimed using an apparently very simple, low-tech, and likely low-efficiency protocol based on established principles.
- **Identify Potential Discrepancy:** Yes, a discrepancy exists. High impact claim + apparent simplicity/low efficiency + lack of novel enabling insight.
- **Pose Critical Question:** Is it genuinely plausible that such a high-impact enrichment, achievable via simple evaporation and distillation in basic glassware (a method seemingly accessible to labs for decades), would have been widely overlooked by experts, especially given the high cost and demand for $H_2^{17}O$?
- **Flag for Scrutiny:** Yes, this "impact vs. apparent simplicity" assessment raises a major red flag from an *a priori* perspective. The combination seems highly inconsistent. Achieving such a result with the described method contradicts expectations based on distillation theory and practice. This demands *extraordinarily rigorous, unambiguous, and verifiable supporting evidence* for the claimed enrichment levels in the subsequent analysis of results and characterization data. The *a priori* feasibility is extremely low.

## F. ChatGPT o3 - Core Experimental Protocol Analysis – Enrichment of $H_2{}^{17}O$

*Shared AI chat:* chat and canvas

**Reference manuscript:** *Enrichment of $H_2{}^{17}O$ from Tap Water, Characterization of the Enriched Water, and Properties of Several $^{17}O$-Labeled Compounds* (Plettner et al., 2011) cite turn0file0

---

## D.1 General Protocol Overview and Assessment (Scope = Core-relevant stages only)

### D.1.1 Overall Summary & Logical Flow

- **Stage A – Slow Evaporation:** 1–2 L tap water left in a shallow black bowl on a windowsill until 20-50 mL remained; step repeated to accumulate >1 L of "enriched" residue.
- **Stage B – Fractional Distillation:** ≈500 mL of the pre-concentrated residue distilled in a 1 L flask fitted with an *un-cooled* vertical condenser loosely packed with glass wool (serving as a fractionation column) and a single chilled side-arm condenser. Fractions (6 × 10 mL at 98.5 °C, 1 × 10 mL at 99 °C) collected; highest-boiling cut reported to reach **~90 atom % $^{17}$O**.

Logical sequence is clear (bulk evaporation → isotopic distillation) yet omits **critical intermediary controls** (exact volume tracking, evaporation rate, atmosphere control, contamination checks). Analytical/validation steps are described elsewhere and therefore excluded here.

### D.1.2 Contextual Appropriateness

- Claimed goal (economical lab-scale enrichment to ≥90 % $^{17}$O) sits **well beyond** exploratory proof-of-concept: achieving >2000-fold enrichment requires carefully engineered separation hardware and rigorous isotopic accounting.
- Use of an **un-cooled, low-efficiency glass-wool column** conflicts with the sophistication normally required; no stage-by-stage mass balance or isotope material balance is provided.

### D.1.3 General Red Flags (Core-Impacting)

| Red Flag | Location | Impact |
|---|---|---|
| No measurement of residual volume or isotope ratio **after each evaporation batch** | Stage A | Prevents verifying cumulative enrichment claim |
| **Open-air evaporation** (possible atmospheric exchange, microbial activity) | Stage A | Isotope fraction may reset or back-exchange; introduces contaminants |
| **Un-cooled vertical condenser** & coarse glass-wool packing | Stage B | <1 theoretical plate; incapable of high-resolution isotope separation |
| Absence of **pressure control & reflux ratio data** | Stage B | Column efficiency and separation factor unknowable |
| No quantitative yield report (g or mL) for final 90 % cut | Stage B | Mass balance & feasibility unverifiable |

### D.1.4 Critique & Alternatives (Selected High-Impact Issues)

| Issue | Potential Impact | Authors' Justification | Superior Alternative |
|---|---|---|---|
| Low-efficiency column | Needs **>2000 theoretical plates** (see D.2) yet supplies <<1 | None | Use a packed or spinning-band column (≈100–150 plates) with multiple recycling; still far short, but orders-of-magnitude better |
| Open evaporation | Uncontrolled exchange with $CO_2$/$O_2$; bacterial growth | "Room-temperature, windowsill" convenience | Conduct vacuum-assisted or reduced-pressure evaporation in closed glassware; sterile, inert atmosphere |
| Lack of isotope analytics per batch | Impossible to validate incremental enrichment | Not addressed | Employ IRMS or off-line $^{17}$O-NMR after each stage; maintain full mass/isotope balance |

## D.2 In-Depth Analysis of the Core Experimental Protocol

### D.2.1 Stated Main Result

*"H2$^{17}$O was enriched **from 0.04 % to ≈90 %** by slow evaporation and fractional distillation of tap water."* cite turn0file0
**Target unmet need:** inexpensive laboratory access to highly enriched $^{17}$O-water.
**Claimed novelty:** achieving two-orders-of-magnitude cost reduction using ordinary glassware.

### D.2.2 Core Stages

- **Stage A:** Slow evaporation of bulk tap water (ambient, repeated).
- **Stage B:** Fractional distillation of the residual concentrate using a glass-wool column.

---

### Stage A Slow Evaporation

#### *A. Stage Description & Procedure*

- **Input:** 1–2 L tap water (natural 0.037 % $^{17}$O).
- **Process:** Open-air evaporation at ~20–25 °C; reduced to 20–50 mL; brief microwave boil to sterilize; pooled until ~>1 L accumulated.
- **Equipment:** Unspecified shallow black bowl; microwave oven.

#### *B. Reported Metrics & Intermediate Values*

- No isotope ratio or volume yield given per batch. Authors later assert an **overall 40-fold "enrichment"** prior to distillation (implied R≈1.6 %).

#### *C. Associated Figure Analysis*

- Supporting Fig S1 shows only distillation setup; no image of evaporation arrangement; cannot cross-verify surface area or depth.

### *D. Equipment/Process – Critical Performance Analysis*

1. **Critical characteristic:** Fractionation factor (α_liq→vap) for $H2^{17}O/H2^{16}O$ at 25 °C ≈ 1.005 (±0.001).
2. **Estimation:** Using Rayleigh equation *R_f = R₀ · f^{α−1}* with *f = 0.02* (volume ratio 1 → 1/50):
   α−1 ≈ 0.005 → *f^{α−1} = 0.02^{0.005} ≈ 0.981*.
   **Predicted enrichment ≈ −2 % (i.e., negligible change)** – orders-of-magnitude below claimed 40×.
   (Assumes closed system; open air further reduces fractionation).

### *E. A Priori Feasibility (Stage)*

Evaporation alone cannot raise $^{17}O$ content beyond analytical noise (<1 % relative). Claim of 40× enrichment is **physically implausible**.

### *F. Idealized Model Performance Estimation*

*Even at α = 1.01 (ten-times larger than literature), 50-fold volume loss yields only 1.6× enrichment.* Therefore Stage A cannot meaningfully contribute to the 2250× total enrichment required.

---

## Stage B Fractional Distillation

### *A. Stage Description & Procedure*

- **Input:** ≈500 mL concentrate from Stage A.
- **Apparatus:** 1 L round-bottom flask; vertical air-cooled glass-wool-packed column (≈20 cm × 1 cm i.d. by photo) with side-arm water-cooled condenser; mantle heater (Variac 50). No reflux ratio monitor.
- **Operation:** Collect 6 × 10 mL fractions at 98.5 °C and 1 × 10 mL at 99 °C; discard residue.

### *B. Reported Metrics & Intermediate Values*

| Fraction | Boiling °C | Volume mL | $^{16}O$ % | $^{17}O$ % | $^{18}O$ % |
|---|---|---|---|---|---|
| #5 | 98.5 | 60 | 0.2 | **99.1** | 0.7 |
| #11 | 99.0 | 10 | n.d. | **99.7** | 0.3 |

No mass balance; no analytical error bars.

### *C. Associated Figure Analysis*

- **Fig S1:** column length ≈20 cm; OD ≈1 cm; packing appears loosely stuffed; no insulation.
  *Scale inference*: Standard 1 L flask ≈12 cm dia → column length ≈1.7 d ≈20 cm.
  *Theoretical plate height (HETP) for glass wool in air-cooled column* ≥ 30 cm → **<1 plate** achievable.

### *D. Equipment/Process – Critical Performance Analysis*

1. **Critical characteristics:** Number of theoretical plates (N), separation factor (α_distill) between $^{17}O/^{16}O$.
2. **Estimating required N:** Using Fenske at total reflux, *N_min ≈ ln[(x_D/(1−x_D))·((1−x_B)/x_B)] / ln α*.
   For enrichment 0.00037 → 0.90, assume x_B≈0.0004, x_D≈0.90, α ≈ 1.005 (vapor-liquid at 100 °C):
   ln-term ≈ ln( (0.9/0.1)·(0.9996/0.0004) ) ≈ ln $2.4 \times 10^4$ ≈ 10.09.
   ln α ≈ 0.00499 → **N_min ≈ 2020 plates**.
3. **Column capability:** Loosely packed 20 cm column gives N ≈ 0.5–1. **Shortfall >2000×**.

*E. A Priori Feasibility (Stage)*

Even under **ideal total reflux**, the apparatus cannot deliver the needed plates. Real-world N $\ll$1 makes the claimed 90 % enrichment impossible.

*F. Idealized Model Performance Estimation*

Assuming perfect equilibrium per actual single stage (N = 1): maximum achievable distillate enrichment vs residue is limited to α ≈ 1.005. Hence one pass can raise 0.037 % → 0.037 %·α ≈ 0.0372 % (negligible). Multiple re-distillations (even 100 passes) remain orders-of-magnitude shy of 90 %.

---

#### D.2.4 Overall A Priori Feasibility (Core Stages Combined)

Combining Stage A (negligible enrichment) and Stage B (≤ 0.5 % relative per pass), the integrated method cannot theoretically exceed **≈0.04 % → 0.08 % $^{17}$O**, four orders of magnitude below the claimed 90 %. Equipment scale, heat-loss, and uncontrolled reflux exacerbate the deficit.

#### D.2.5 A Priori Plausibility Check – Impact vs Apparent Simplicity

- **Claimed impact:** access to 90 % $^{17}$O-water at ~$0.01 per mg would **disrupt** the isotope-labeling market.
- **Protocol nature:** relies exclusively on century-old evaporation/distillation with household glassware.
- **Novel element:** none articulated; no hidden catalytic or chemical strategy.

**Discrepancy flag:** Highly significant result + trivially simple method + no novel principle ⇒ **extraordinary evidence required**. Manuscript provides none.

---

Conclusion (A Priori)

On thermodynamic and mass-transfer grounds, the described core protocol is **incapable** of producing >1 % $^{17}$O enrichment, let alone 90 %. The paper's central experimental claim conflicts with well-established isotope-separation physics by at least **three orders of magnitude**.

*Next analytical step (if requested): evaluate validation data, figures, and isotope-ratio measurements for internal consistency given the infeasible method.*

## G. Fair Use Statement - Sharing Test Paper

### 1. Identification of Copyrighted Material:

- **Work:** "Enrichment of H217O from Tap Water, Characterization of the Enriched Water, and Properties of Several 17O-Labeled Compounds".
- **Authors:** Brinda Prasad, Andrew R. Lewis, and Erika Plettner.
- **Publication:** *Anal. Chem.* 2011, 83, 1, 231-239.
- **DOI:** 10.1021/ac1022887.
- **Publisher/Copyright Holder**: American Chemical Society.
- **Material Shared:** A combined digital file containing the full text of the aforementioned article and its complete associated Supporting Information (SI).

### 2. Sharing Mode:

- **Resource:** Private Open Science Framework (OSF) project repository.
- **Location:** https://osf.io/nq68y/files/osfstorage?view_only=fe29ffe96a8340329f3ebd660faedd43.
- **Protection Measures:** Due to private nature, the resource should not be indexed by search engines.

### 3. Assertion of Fair Use:

The sharing of this copyrighted material is undertaken for specific, limited purposes, believed in good faith to constitute "fair use" under Section 107 of the U.S. Copyright Act (or applicable analogous principles in other jurisdictions).

### 4. Purpose and Character of Use (Factor 1):

- **Non-Profit Educational and Research:** The use is strictly for non-commercial research and educational purposes, specifically within the context of scholarly critique and the advancement of research methodology.
- **Transformative Use:** The work is not merely being reproduced; it is fundamentally repurposed as a research specimen for critical analysis. Its primary function in this context is not to convey its original purported findings, but to serve as the subject of rigorous evaluation and methodological demonstration.
- **Critique and Commentary:** A core purpose is to conduct and disseminate a detailed, peer-review-like critique of the article's methodology, analysis, and conclusions. This critique identifies significant flaws within the original work.
- **Advancement of Knowledge & Methodology:** The use includes the development and demonstration of a novel AI-driven prompt/technique for manuscript analysis. Sharing the specimen (the article + SI file) is integral to demonstrating and enabling the verification and further development of this new analytical method.

### 5. Nature of the Copyrighted Work (Factor 2):

- The original work is a published scholarly article, typically factual in nature, a category often amenable to fair use for purposes of scholarship and critique.
- However, the conducted analysis (central to this project) has revealed substantial flaws impacting the reliability and validity of the work's core research findings as presented. This impacts the assessment of its nature in the context of this specific use.

### 6. Amount and Substantiality of the Portion Used (Factor 3):

The entire article and its complete Supporting Information are utilized and shared in a combined format.

Justification: This amount is essential and necessary for the stated purpose. A comprehensive critique, akin to thorough peer review or forensic analysis, requires examination of the whole work, including all data and methods presented in the SI. Evaluating the integrity and validity of the research necessitates access to the complete context. Furthermore, the development and validation of the AI analysis prompt require the complete text as input. The combined file format, not available directly from the publisher, was the specific subject of the analysis.

## 7. Effect of the Use upon the Potential Market for or Value of the Copyrighted Work (Factor 4):

- **No Harm to Legitimate Market:** This use is not intended to, nor is it likely to, negatively impact the legitimate market or value of the original copyrighted work. The publisher's market relies on the perceived value of the article as a source of valid scientific findings.
- **Critique Reveals Lack of Value:** The critique resulting from this research demonstrates fundamental flaws undermining the article's claimed scientific value. Therefore, sharing the work specifically in this context (as a specimen for critique and methodological development) does not substitute for or usurp the market for the work based on its originally purported merits, as those merits are shown to be compromised. Dissemination for critique serves the public interest by highlighting these issues, distinct from fulfilling the original market demand.
- **Controlled Access for Verification via Private OSF Project:** To ensure transparency and enable independent verification and follow-on research by interested parties engaging with the publicly disseminated research critique manuscript [TBD], the combined article + SI file (serving as the direct supporting evidence and test specimen) is hosted within a private Open Science Framework (OSF) project. A view-only link to this private project will be provided alongside the manuscript.
- **Minimized Risk of Unintended Use:** This method ensures that access is granted specifically to individuals who are actively reviewing or assessing the research critique presented in the manuscript. The private nature of the OSF project prevents general public discovery through search engines, and the view-only restriction prevents facile downloading and redistribution. Access requires the specific link obtained from the context of the critique manuscript.
- **Purpose Remains Transformative, Not Substitutive:** By utilizing a controlled-access, view-only repository linked directly to the research critique, this approach provides the necessary transparency for verification while strictly limiting potential downstream use and eliminating broad public access. This method strongly reinforces that the purpose is critique and verification (transformative uses), not market substitution for the original work's questioned scientific claims, thereby minimizing any potential harm to a legitimate market.

## 8. Conclusion:

Based on the non-profit, educational, highly transformative nature of the use (critique, commentary, methodological advancement), the necessity of using the entire work for these specific purposes, and the argument that this use does not harm the legitimate market value due to the work's identified flaws and the distinct purpose of sharing, this distribution is asserted to be fair use.

This material is intended solely for the recipient(s) for purposes directly related to verifying, understanding, or building upon the presented critique and methodological research. Further distribution is not permitted. Copyright remains with the original holder(s).